\documentclass[sn-mathphys-num]{sn-jnl}

\usepackage{graphicx}%
\usepackage{multirow}%
\usepackage{amsmath,amssymb,amsfonts}%
\usepackage{amsthm}%
\usepackage{mathrsfs}%
\usepackage[title]{appendix}%
\usepackage{xcolor}%
\usepackage{textcomp}%
\usepackage{manyfoot}%
\usepackage{algorithm}%
\usepackage{algorithmicx}%
\usepackage{algpseudocode}%
\usepackage{listings}%
\usepackage{epsfig}
\usepackage{subfigure}
\usepackage{caption}
\usepackage{soul,color}
\usepackage{booktabs}
\usepackage{CJK}
\usepackage{makecell}
\usepackage{wrapfig}
\usepackage{picinpar}
\usepackage{cutwin}
\usepackage{colortbl}

\theoremstyle{thmstyleone}%
\theoremstyle{thmstyletwo}%

\theoremstyle{thmstylethree}%

\begin{document}

\title[Article Title]{Part-Whole Relational Fusion Towards Multi-Modal Scene Understanding}


\author[1]{\fnm{Yi} \sur{Liu}}\email{liuyi0089@gmail.com}

\author[1]{\fnm{Chengxin} \sur{Li}}\email{s23150812015@smail.cczu.edu.cn}

\author[1]{\fnm{Shoukun} \sur{Xu}}\email{jpuxsk@163.com}
\equalcont{Equally corresponding authors.}
\author[2]{\fnm{Jungong} \sur{Han}}\email{jungonghan77@gmail.com}
\equalcont{Equally corresponding authors.}

\affil[1]{\orgdiv{School of Computer Science and Artificial Intelligence}, \orgname{Changzhou University}, \city{Changzhou}, \postcode{213000}, \state{Jiangsu}, \country{China}}

\affil[2]{\orgdiv{Department of Computer Science}, \orgname{University of Sheffield}, \city{Sheffield}, \postcode{S10 2TN}, \country{U.K}}



\abstract{Multi-modal fusion has played a vital role in multi-modal scene understanding. Most existing methods focus on cross-modal fusion involving two modalities, often overlooking more complex multi-modal fusion, which is essential for real-world applications like autonomous driving, where visible, depth, event, LiDAR, etc., are used. Besides, few attempts for multi-modal fusion, \emph{e.g.}, simple concatenation, cross-modal attention, and token selection, cannot well dig into the intrinsic shared and specific details of multiple modalities. To tackle the challenge, in this paper, we propose a Part-Whole Relational Fusion (PWRF) framework. For the first time, this framework treats multi-modal fusion as part-whole relational fusion. It routes multiple individual part-level modalities to a fused whole-level modality using the part-whole relational routing ability of Capsule Networks (CapsNets). Through this part-whole routing, our PWRF generates modal-shared and modal-specific semantics from the whole-level modal capsules and the routing coefficients, respectively. On top of that, modal-shared and modal-specific details can be employed to solve the issue of multi-modal scene understanding, including synthetic multi-modal segmentation and visible-depth-thermal salient object detection in this paper. Experiments on several datasets demonstrate the superiority of the proposed PWRF framework for multi-modal scene understanding. The source code has been released on {\color{blue}https://github.com/liuyi1989/PWRF}.}




\maketitle

\section{Introduction}
\label{sec:Introduction}
Due to the limited perception ability of single sensor, multiple sensors help to capture different fields of perception, \emph{e.g.}, depth, thermal, and LiDAR \cite{huang2021makes, wang2023mul, liu2024coc, planamente2024cre, zhu2024uni}. Naturally, multi-modal fusion plays a fundamental role in multi-sensor scene understanding, with applications ranging from synthetic autonomous driving perception \cite{cao2021shapeconv, wang2023mul} to unmanned aerial vehicles \cite{sun2022drone}. 

Previous multi-modal fusion methods mostly lie in the cross-modal combinations \cite{chen2021spatial, xiang2021polarization, zhou2021gmnet}, which aim to find the complementary details in different modalities to enhance individual representations. Despite these methods have advanced the progress of multi-modal fusion, they largely concentrate on specific sensor pairs and lack behind the current trend of fusing multiple modalities \cite{wang2022multimodal}. In contrast, the noise and misalignment of multi-modal sensors pose significant challenges for fusing more modalities, such as triple-modal fusion, which is crucial for many autonomous driving applications. 
\begin{figure}[htbp]
	\centering
	\includegraphics[width=0.92\linewidth]{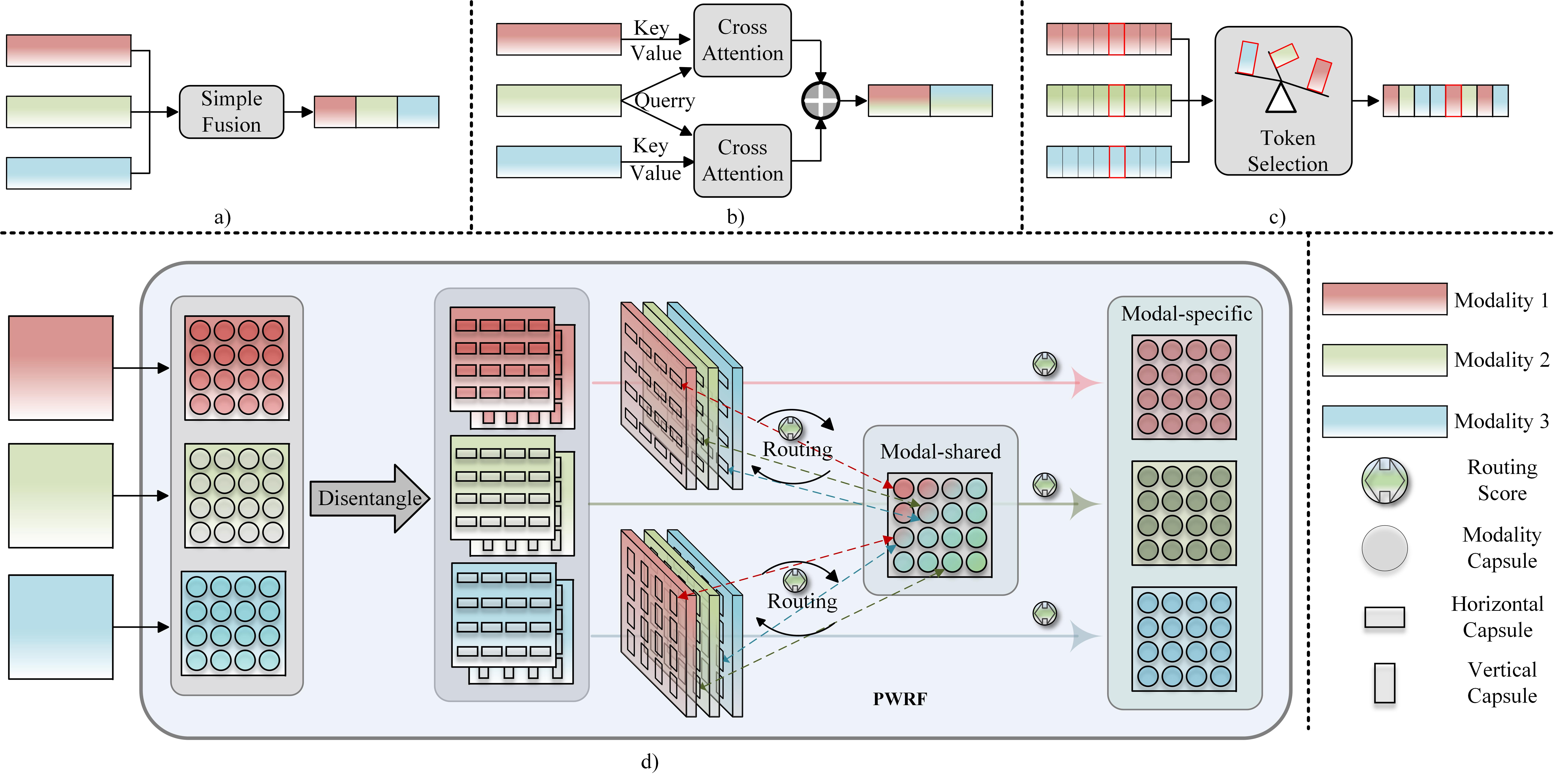}
	\caption{Comparison of different multi-modal fusion methods. (a) Multi-modal fusion via concatenation. (b) Multi-modal fusion through parallel cross-attention to attend the primary modality. (c) Multi-modal fusion via selection mechanism. (d) Our multi-modal fusion via part-whole relational routing to generate modal-shared and modal-specific details.}
	\label{fig:introdution}
\end{figure}

Within the scope of multi-modal fusion\footnote{In this paper, multi-modal fusion focuses on triple-modal fusion instead of cross-modal fusion.}, there are a few approaches towards this field, as shown in Fig. \ref{fig:introdution}. For example, Fig. \ref{fig:introdution}(a) explores the triple-modal fusion using a simple concatenation \cite{wan2023mffnet}. Fig. \ref{fig:introdution}(b) uses parallel cross-attention mechanism to attend the primary modality \cite{broedermann2023hrfuser}. Fig. \ref{fig:introdution}(c) selects the informative-modal token from three modalities as the fusion result \cite{wang2022multimodal, zhang2023delivering}. To sum up, the existing multi-modal fusion approaches mostly combine the important details from multiple modalities using attention \cite{broedermann2023hrfuser, wang2022multimodal} or select the most important modality for each patch using a maximum criteria \cite{zhang2023delivering}. Albeit some progress, these methods still encounter challenges for arbitrary-modal fusion. First, attention-based fusion methods \cite{broedermann2023hrfuser, wang2022multimodal} focus on identifying important details rather than capturing intrinsic knowledge among multiple modalities, including shared and specific knowledge, thus lacking intrinsic fusion effectiveness. Secondly, some methods discard many supplementary modalities, \emph{e.g.}, the informative modal selection method \cite{zhang2023delivering}, which can cause performance degradation in scenarios requiring the fusion of more modalities.

To tackle the issue of the multi-modal fusion, we explore an alternative fusion, called Part-Whole Relational Fusion (PWRF), which treats the relationships between individual modality and the fusion modality as relationships between each part-level modality and the whole-level modality. In such sense, the solution of multi-modal fusion involves routing part-modalities to the whole-modality, which can be achieved through the part-whole relational routing within the framework of Capsule Network (CapsNets) \cite{hinton2018matrix}. To this end, considering the heavy computation of CapsNets, we opt for a lightweight version \cite{liu2022disentangled}, named Disentangled Capsule Routing (DCR), to routing the part-modality to whole-modality. Concretely, DCR begins by disentangling part-level modal capsules from each single modality along the horizontal and vertical dimensions, which are fed into capsule routing mechanism to generate horizontal and vertical whole-level modal capsules. These orthogonal capsules are entangled to achieve the whole-level modal capsules. Thanks to the primitive fusion of PWRF in Fig. \ref{fig:introdution}(d), modal-shared and modal-specific semantics, which are two vital fusion semantics in the issue of multi-modal fusion \cite{wang2023multi}, are included. To be concrete, modal-shared semantics are represented by the whole-level modal capsules, generated by exploring common properties across different individual modalities. Modal-specific semantics for each modality are computed via the routing coefficients from each part-level modality to the whole-level modality, reflecting the associations between each single modality and the fused version.

To explore the potential of the proposed PWRF for multi-modal scene understanding, we select two fundamental tasks to validate its superiority: Synthetic Multi-Modal (SMM) semantic segmentation \cite{zhang2023delivering} and Visible-Depth-Thermal (VDT) salient object detection \cite{wan2023mffnet}. Synthetic Multi-Modal (SMM) semantic segmentation \cite{zhang2023delivering} and Visible-Depth-Thermal (VDT) salient object detection \cite{wan2023mffnet} are chosen due to their complementary focus areas in multi-modal scene understanding. Specifically, SMM semantic segmentation is concerned with segmenting various semantic classes in multi-modal input, which includes visible, depth, event, and LiDAR data, thereby enhancing detailed scene understanding by leveraging the strengths of each modality for comprehensive semantic representation. In contrast, VDT salient object detection is a crucial task in autonomous driving and robotic navigation, where the primary focus is to identify the most salient objects in challenging multi-modal settings involving visible, depth, and thermal data. This task is essential for applications that prioritize real-time decision-making, especially under adverse or uncertain conditions where identifying key objects becomes critical for effective scene understanding. Inspired by the above concerns, these two tasks provide a holistic evaluation of the proposed PWRF framework for the issue of multi-modal scene understanding. Concretely, SMM segmentation assesses the capability of the framework to understand and represent detailed semantic aspects of the scene, while VDT saliency detection focuses on the ability to extract and prioritize critical information. The integration of both tasks showcases the versatility of our approach in tackling different facets of multi-modal scene understanding, thus demonstrating its generalizability and robustness across varied application scenarios. Experiments on SMM semantic segmentation \cite{zhang2023delivering} and VDT salient object detection \cite{wan2023mffnet} datasets prove the superiority of the proposed PWRF framework for multi-modal scene understanding.

Contributions of this paper are described as follows:

(i) We propose a PWRF framework for multi-modal scene understanding, which, to the best of our knowledge, is the first to treat multi-modal fusion as the part-whole relational fusion. Under this framework, we can obtain modal-shared and modal-specific details using the whole-level modality and routing coefficients, respectively, which can be further employed to enhance multi-modal scene understanding.

(ii) To apply PWRF for SSM semantic segmentation, we design a shared-specific-integration module that fuses modal-shared and modal-specific details to detect semantic objects in synthetic multi-modal scenarios.

(iii) In order to configure PWRF for VDT salient object detection, we design a stacking adjacent-scale attention decoder to integrate modal-shared and modal-specific details, which is experimentally proved to be superior to detect the salient objects in the visible-depth-thermal environment.

The paper is organized as follows. Sec. \ref{sec:Related} reviews the works related to the proposed method. Sec. \ref{sec:Proposed} details the proposed PWRF framework. Sec. \ref{Multi-modal-scene} illustrates the architectures for SMM semantic segmentation and VDT salient object detection using PWRF. Sec. \ref{sec:Experiment} carries out abundant experiments and analyses to understand the proposed method. Sec. \ref{sec:Conclusion} provides a conclusion for the paper.

\section{Related work}
\label{sec:Related}
In this section, we will review the works related to our method, including multi-modal semantic segmentation, multi-modal salient object detection, and CapsNets, which will be described in detail in the following.

\subsection{Multi-modal semantic segmentation}
Semantic segmentation has been a fundamental task in the computer vision community since fully convolutional networks \cite{long2015fully} revolutionized its development. Unlike RGB-modal semantic segmentation \cite{jin2021mining, borse2021inverseform, gu2022multi} that relies on RGB modality, which may suffer from sensor limitations, multi-modal semantic segmentation enhances scene perception by incorporating multiple modalities, \emph{e.g.}, depth, events, and LiDAR. For example, Hazirbas \emph{et al.} \cite{hazirbas2017fusenet} leveraged the rich color and texture from RGB modality, and geometric and structural information from depth modality, for semantic segmentation. Wang \emph{et al.} \cite{wang2016learning} proposed to extract common and modality-specific features from RGB and depth images to improve semantic segmentation accuracy for indoor scenes. Wang \emph{et al.} \cite{wang2022multimodal} proposed a multi-modal token fusion method through substituting uninformatie tokens with important ones. Wang \emph{et al.} \cite{wang2020deep} dynamically exchanged channels between sub-networks of different modalities for multi-modal fusion. Zhao \emph{et al.} \cite{zhao2023lif} presented a coarse-to-fine fusion mechanism for LiDAR-camera semantic segmentation by leveraging the low-level contextual information and designing an offset correction. Liu \emph{et al.} \cite{liu2022camliflow} fused camera and LiDAR modalities in a bi-directional manner. Liang \emph{et al.} \cite{liang2022multimodal} derived a region guided filter to select informative combinations of multiple modalities classes. Zhang \emph{et al.} \cite{zhang2023cmx} utilized a cross-modal feature correction module to enhance complementary information and a feature fusion module to achieve full fusion and long-distance context exchange. Based on this, they subsequently presented a cross-modal segmentation model \cite{zhang2023delivering} by fusing the primary modality and selected informative auxiliary modality. 

Most of the previous methods focus on fusing complementary cues \cite{wang2022multimodal} or selecting informative one modality while discarding the others, which can lead to performance degradation due to a lack of intrinsic integration. Differently, we treat multi-modal fusion as part-whole relational fusion that routes part-level modalities to whole-level modality, thus enabling the capture of primitive fusion and generating shared and specific details.
\subsection{Multi-modal salient object detection}
Unlike RGB salient object detection \cite{tian2023modeling, li2023salient, yuan2023ctif, liu2021part, liu2023tcgnet}, which often severely struggles challenging scenarios such as cluttered background and high similarity between salient objects and their surroundings, multi-modal learning can enhance saliency understanding in these difficult conditions. The most popular multi-modal salient object detection lies in RGB-D and RGB-T saliency, which detect the salient objects from the RGB \& depth and RGB \& thermal data, respectively, via cross-modal fusion. For example, Wu \emph{et al.} \cite{wu2023hidanet} proposed a multi-level and multi-scale fusion scheme to fuse RGB-depth features for saliency prediction. Li \emph{et al.} \cite{li2023delving} designed a boundary-aware fusion framework for RGB-D salient object detection. Xie \emph{et al.} \cite{xie2023cross} developed a double bi-directional interaction network for RGB-D saliency detection. Zhang \emph{et al.} \cite{zhang2023saliency} utilized saliency prototypes of the primary modality to enhance semantics of the auxiliary modality, followed by allocating dynamically weighs for auxiliary modality during fusion stage. Different from RGB-D and RGB-T salient object detection, the VDT salient object detection tackles the problem of salient object in the environment of the visible, depth, and thermal images. To solve the challenge, Song \emph{et al.} \cite{song2022novel} utilized an attention mechanism to parallelly fuse the primary modality and the auxiliary modality for triple-modal salient object detection. This approach highlights an innovative method for complementary aggregation of triple-modal information. Wan \emph{et al.} \cite{wan2023mffnet} proposed a triple-modal fusion encoder and a progressive feature enhancement decoder for VDT salient object detection. They further designed a triple-modal interaction encoder and a multi-scale fusion decoder for VDT salient object detection \cite{wan2024tmnet}. Bao \emph{et al.} \cite{bao2024quality} developed a quality-aware selective fusion network for VDT saliency detection. 

Different from the existing VDT salient object detection that cannot explore the modal-shared and modal-specific semantics, our PWRF framework can explore the associations between triple modalities to find modal-shared and modal-specific semantics for further saliency prediction.
\subsection{CapsNets}
Unlike convolutional neural networks that intend to capture the discriminative features, CapsNets target at capturing part-whole relations to find the targets. The classical CapsNets, \emph{i.e.}, vector CapsNets \cite{sabour2017dynamic} and matrix CapsNets \cite{hinton2018matrix}, have been known large-scale parameters and heavy computation. To achieve lightweight CapsNets, a lot of efforts have been devoted. For example, a prediction tuning framework was proposed to allow a deep architecture \cite{pan2021pt}. Shi \emph{et al.} \cite{shi2022sparse} utilized sparse optimization to compress CapsNets via reducing unnecessary weight parameters and computational cost. In our previous work, a residual pose routing \cite{liu2024capsule} and a disentangled-entangled routing mechanism \cite{liu2022disentangled} were proposed to speed up CapsNets. In addition to CapsNets architectures, they have been introduced to a wide range of applications. Jampour \emph{et al.} \cite{jampour2021capsnet} introduced a new regularization term into CapsNets to improve the generalization for signature identification.Our previous works employed CapsNets for part-whole relational visual saliency \cite{liu2021part} and visual camouflage \cite{liu2021integrating}. Wu \emph{et al.} \cite{wu2022interpretable} fed capsule features from multiple modalities into long short-term memory for motion recognition. 

In this paper, we apply CapsNets for multi-modal scene understanding by the proposed PWRF framework. Different from \cite{wu2022interpretable} that utilized CapsNets to extract modal capsules features instead of multi-modal fusion, we employ the part-whole relational routing ability of CapsNets for multi-modal fusion stage.
\section{Proposed Part-Whole Relational Fusion}
\label{sec:Proposed}

In this section, we will describe the details of PWRF, which takes multi-modal fusion as routing multiple individual part-level modalities to the fused whole-level modality. The framework PWRF consists of two core components, including part-whole modality routing, and modal-shared and modal-specific details generation. We take triple modalities as the case for illustration in the following.

\subsection{Part-whole modality routing}


To achieve primitive fusion of multiple modalities, we take each single modality and the fused modality as a part-level modality and the whole-level modality, respectively, which guides multi-modal fusion to the issue of routing multiple part-level modalities to their whole-level modality. Leveraging the part-whole relational routing ability of CapsNets \cite{hinton2018matrix}, we implement the part-whole relational fusion. However, considering the heavy computation demands of the original CapsNets \cite{hinton2018matrix}, we use our previous lightweight capsule routing version, called Disentangled Capsule Routing (DCR) algorithm \cite{liu2022disentangled}, as an alternative. In the following, we detail the PWRF framework using DCR for routing part-level modalities to the whole-level modality, which is composed by part-level modality capsules disentanglement and part-level to whole-level capsules routing.

\subsubsection{Part-level modality capsules disentanglement}
Part-level modality capsules disentanglement targets at disentangling the full-resolution capsule features of each single modality to two orthogonal versions, including horizontal and vertical capsules, which will transform two-dimensional capsule routing to one-dimensional version with computation reduced greatly. 

Given features $\left \{ {\bf{f}}_i^n \in {\Re ^{H_i \times W_i \times C_i}} \mid  n\in \left \{1, 2, 3  \right \} ,i=1, 2, 3, 4\right \} $ from the backbone network of ResNet-50 \cite{he2016deep}, where $H_i \times W_i$ and $C_i$ represent the resolution and channel of features at stage-$i$ for modality $n$, respectively. Each single modality capsule, including pose matrix ${\bf{p}}_i^n$ and activation value $\boldsymbol{a}_i^n$, can be constructed as follows:
\begin{equation}
	\label{equ:Primarycapsules}
	\begin{array}{l}
		{\bf{p}}_i^n \in {\Re ^{{H_i} \times {W_i} \times T^p \times 16}} = Conv\left( {{\bf{f}}_i^n}, {dim}=3 \right),\\
	    \boldsymbol{a}_i^n \in {\Re ^{{H_i} \times {W_i} \times T^p \times 1}} = \sigma (Conv\left( {{\bf{f}}_i^n}, {dim}=3 \right)),
	\end{array}
\end{equation}
{where $Conv\left( { * ,dim = 3} \right)$ and {$\sigma$} refer to convolution and {Sigmoid} operations along the 3${rd}$ channel, respectively.} $T^p$ is the type number of the part-level capsules. 16 and 1 represents the pose matrix dimension and activation value dimension, respectively. {The full-resolution capsules ${\mathbf{P}_{i}^n}$ is composed by 
\begin{equation}
	\label{equ:Primarycapsulesfull}
	{\mathbf{FP}_{i}^n \in \Re^{H_i \times W_i \times T^p \times 17} = {Concat}(\mathbf{p}_i^n, \boldsymbol{a}_i^n, {dim} = 4)}.
\end{equation}

{On top of the full-resolution capsules ${\mathbf{P}_{i}^n}$, we disentangle the horizontal 1-dimensional capsules as follows}
\begin{equation}
	\label{equ:hor-disentangle}
	{\mathbf{P}_{i,H}^n \in \Re^{H_i \times 1 \times T^p \times 17} = {Conv}(\mathbf{FP}_i^n, {dim} = 2)}.
\end{equation}

Similarly, the vertical part-level capsules can be disentangled as
\begin{equation}
	\label{equ:ver-disentangle}
	{\mathbf{P}_{i,V}^n \in \Re^{1 \times W_i \times T^p \times 17} = {Conv}(\mathbf{FP}_i^n, {dim} = 1)}.
\end{equation}
\subsubsection{Part-level to whole-level capsules routing}
{The disentangled horizontal and vertical part-level modal capsules, ${\mathbf{P}_{i,H}^n}$ and ${\mathbf{P}_{i,V}^n}$, will be fed into the capsule routing algorithm \cite{hinton2018matrix} to find the whole-level modal capsules via exploring part-whole relations.}
Specifically, part-level capsules of multiple modalities are combined together, which is achieved by concatenating horizontally part-level capsules of multiple modalities and vertically part-level capsules of multiple modalities along the type dimension separately, \emph{i.e.},
\begin{equation}
	\label{equ:cap-h-concat}
	\begin{array}{l}
		\mathbf{P}_{i,H} \in {\Re ^{{H_i} \times 1 \times ({T^p} + {T^p} + {T^p}) \times 17}} = Concat\left( {{{\left. {\mathbf{P}_{i,H}^n} \right|}_{n = 1,2,3}},dim  = 3} \right)
	\end{array},
\end{equation}
\begin{equation}
	\label{equ:cap-v-concat}
	\begin{array}{l}
	\mathbf{P}_{i,V} \in {\Re ^{1 \times {W_i} \times ({T^p} + {T^p} + {T^p}) \times 17}} = Concat\left( {{{\left. {\mathbf{P}_{i,V}^n} \right|}_{n = 1,2,3}},dim  = 3} \right)
	\end{array},
\end{equation}

The horizontally and vertically whole-level modal capsules will be computed by implementing capsule routing \cite{hinton2018matrix} on horizontally and vertically part-level modal capsules separately as follows

\begin{equation}
	\label{equ: h-routing}
	\begin{array}{l}
		{\mathbf{W}_{i,H} \in \Re^{H_i \times 1 \times T^w \times 17}, \mathbf{R}_{i,H} \in {\Re ^{{ ({H_i} \times 1)\times (T^p + T^p + T^p)} \times {T^w}}} = Routing\left( \mathbf{P}_{i,H} \right)}
	\end{array},
\end{equation}
\begin{equation}
	\label{equ: v-routing}
	\begin{array}{l}
		{ \mathbf{W}_{i,V} \in \Re^{1 \times W_i \times T^w \times 17}, {\mathbf{R}_{i,V}} \in {\Re ^{{(1 \times {W_i}) \times (T^p + T^p + T^p)} \times {T^w}}} = Routing\left(  \mathbf{P}_{i,V} \right)}
	\end{array},
\end{equation}
{where $Routing\left(  *  \right)$ represents the capsule routing algorithm \cite{hinton2018matrix}. {$\mathbf{R}_{i,H}$ and $\mathbf{R}_{i,V}$} are the routing coefficients from part-level modalities to the whole-level modal, which reveal the familiar relations between the part-level modalities and the whole-level modality along the horizontal and vertical dimensions, respectively. ${\mathbf{W}_{i,H}}$ and ${\mathbf{W}_{i,H}}$ are the horizontal and vertical whole-level modalities, respectively.}

On top of that, the full-resolution whole-level modal capsules can be achieved by entangling them as follows
\begin{equation}
	\label{equ:shd-cap-disentangle}
	{\mathbf{WP}_{i} \in {\Re ^{{H_i} \times {W_i} \times {T^w} \times 17}} = \mathbf{W}_{i,H} \otimes {\mathbf{W}_{i,V} },}
\end{equation}
where $ \otimes $ represents the operation of matrix multiplication along the resolution dimension. 
\subsection{Modal-shared and modal-specific details generation}

\subsubsection{Modal-shared details} 
{The whole-level modality ${{\bf{WP}}_i}$ in Eq. (\ref{equ:shd-cap-disentangle}) captures associations across different individual modalities, since it is computed by routing information from multiple modalities. The generation of modal-shared details is a crucial step in the PWRF framework, allowing the model to leverage information that is consistent across multiple modalities. The modal-shared details are generated by aggregating features from all part-level modalities through a part-whole relational routing mechanism. In light of this fact, we treat it as the modal-shared semantic details.}

\subsubsection{Modal-specific details}
The routing coefficients $\mathbf{R}_{i,H}$ and $\mathbf{R}_{i,V}$ in Eqs. (\ref{equ: h-routing}) and (\ref{equ: v-routing}) indicate the likelihood of each part-level capsule belonging to each latent whole-level capsule class, which reveal the contributions of different single part-level modalities to the whole-level modality, respectively. Therefore, we model the modal-shared details for each modality using the routing coefficients and each modalities capsules. First, the part-level correlations for each modality can be extracted from the horizontal and vertical routing coefficients $\text{R}_{i,H}$ and $\text{R}_{i,V}$

\begin{equation}
	\label{equ:spc-hc}
	{\text{R}_{i,H}^n  \in \Re^{H_i \times 1 \times T^p} = \frac{1}{T^w} \sum_{k=1}^{T^w} \mathbf{R}_{i,H}[:,:,T^p(n-1)+1:T^p n,k], \quad n = 1, 2, 3, }
\end{equation}
\begin{equation}
	\label{equ:spc-vc}
	{\text{R}_{i,V}^n \in \Re^{1 \times W_i \times T^p} = \frac{1}{T^w} \sum_{k=1}^{T^w} \mathbf{R}_{i,V}[:,:,T^p(n-1)+1:T^p n,k], \quad n = 1, 2, 3,}
\end{equation}
{where \(\text{R}_{i,H}^n\) and \(\text{R}_{i,V}^n\) represent the horizontal and vertical routing coefficients for each part of the capsule corresponding to $n^{\text{th}}$ modality, respectively. The third dimension of the routing coefficients is split into three parts, each of size \(T^p\).}
\begin{figure*}[htbp]
	\centering
	\includegraphics[width=0.92\linewidth]{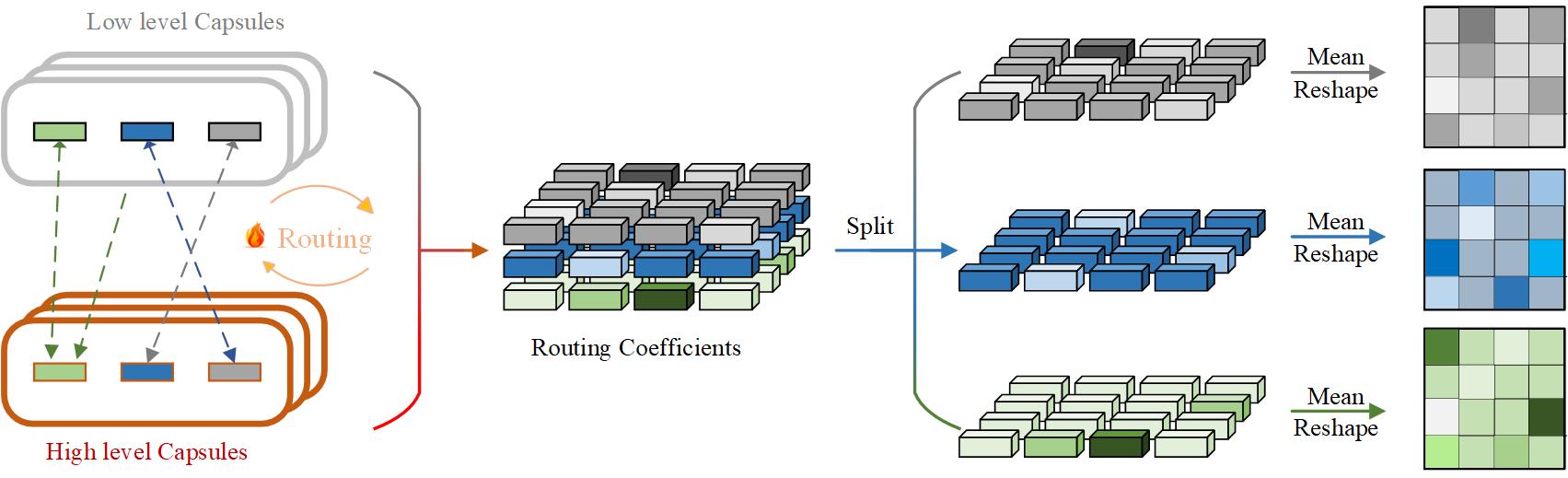}
	\caption{{Visualization for the split of routing coefficients.}}
	\label{fig:rsplit}
\end{figure*}

The horizontal and vertical modal-specific details can be computed via multiplying the each part-level correlations and its capsule features
\begin{equation}
	\label{equ:spc-cap-h}
	{{\bf{SP}}_{i,H}^n \in {\Re ^{{H_i} \times 1 \times {T^p} \times 17}} = { \mathbf{P}_{i,H}^n \odot \text{R}_{i,H}^n},}
\end{equation}
\begin{equation}
	\label{equ:spc-cap-v}
	{{\bf{SP}}_{i,V}^n \in {\Re ^{1 \times {W_i} \times {T^p} \times 17}} = { \mathbf{P}_{i,V}^n \odot \text{R}_{i,V}^n}},
\end{equation}
where ${ {  \odot  } }$ means element-wise multiplication. {The full-resolution modal-specific details are further computed by entangling ${\bf{SP}}_{i,H}^n$ and ${\bf{SP}}_{i,V}^n$ along the resolution dimension as follows}
\begin{equation}
	\label{equ:spc-cap-disentangle}
	{{\bf{SP}}_i^n \in {\Re ^{{H_i} \times {W_i} \times {T^p} \times 17}} = {\bf{SP}}_{i,H}^n \otimes {\bf{SP}}_{i,V}^n.}
\end{equation}
{It is noted that each modality specific details ${\bf{SP}}_i^n$ is reshaped to ${\bf{SP}}_i^n\in {\Re ^{{H_i} \times {W_i} \times ({T^p} \times 17)}}$ by merging the last two dimensions together for the following utilization. On top of that, modal-specific components ${\bf{SP}}_i^n$ of multiple modalities are integrated to get the merged modal-specific details as}
\begin{equation}
	\label{mu}
	{{{\bf{SP}}_i} = ConvCate\left( ({\bf{SP}}_i^n), dim=3 \right),}
\end{equation}
where $ ConvCate(*) $ denotes the operation that combines concatenation and convolution operation along the specified dimension.

{To make the entire routing process more intuitive, Fig. \ref{fig:rsplit} visualizes the  the process of spliting the routing coefficients. As shown in Fig. \ref{fig:rsplit}, on top of the routing between the high-level capsules and the low-level capsules, the routing coefficients are obtained. In order to obtain the corresponding routing coefficients for the respective modalities, the routing coefficients are split and averaged along the last dimension. The reshape operation is gone through for subsequent processing.}
\section{Multi-modal scene understanding using PWRF}
\label{Multi-modal-scene}

In this section, we select two fundamental tasks, including SMM semantic segmentation and VDT salient object detection, to explore the contributions of our PWRF framework for multi-modal scene understanding.
\subsection{SMM semantic segmentation}
\begin{figure*}[htbp]
	\centering
	\includegraphics[width=1\linewidth]{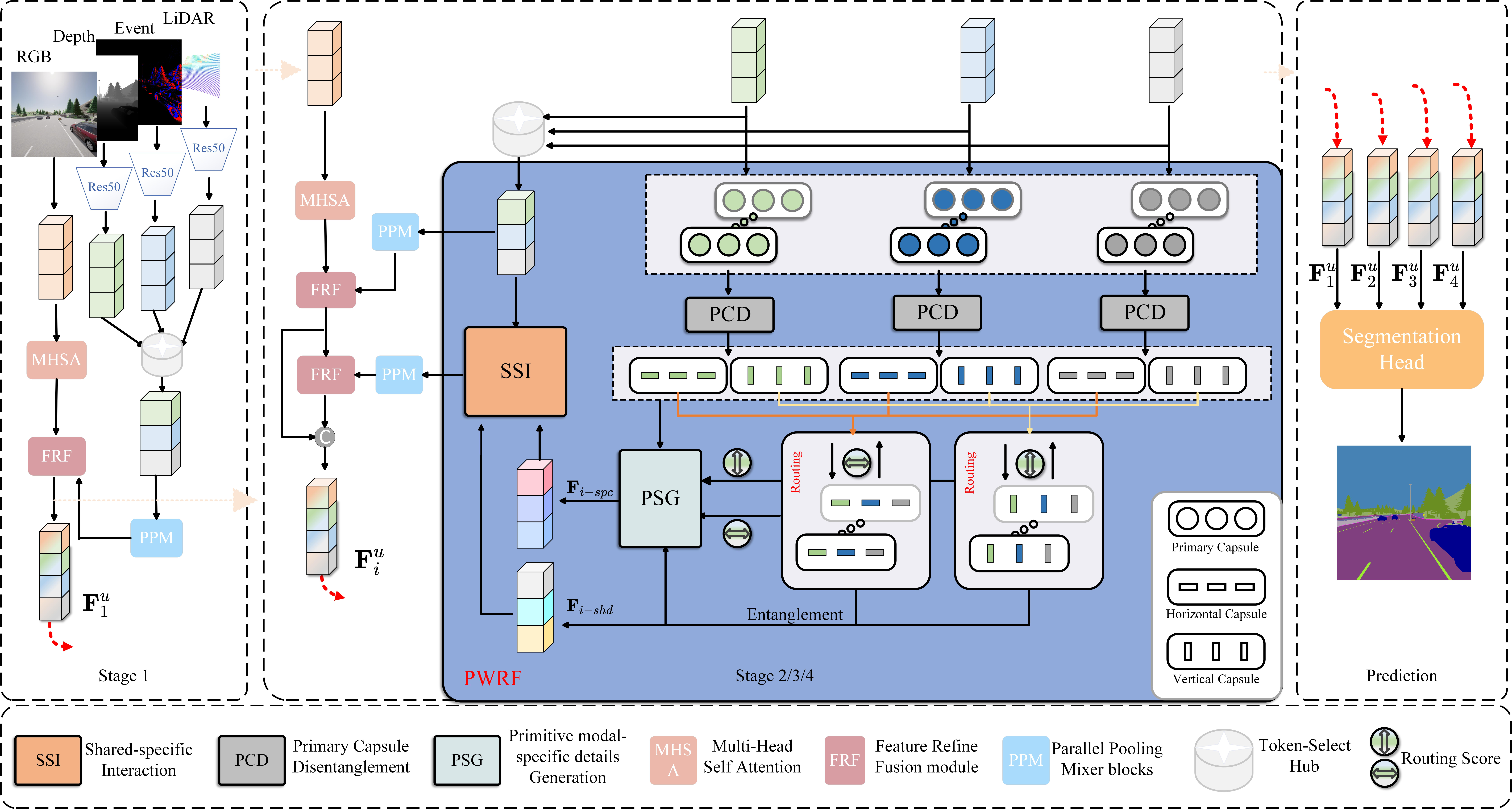}
	\caption{SMM semantic segmentation framework based on PWRF. There are 4 stages with different-scale features and outputs. It is noted that we use the same stage 1 as in \cite{zhang2023delivering} due to the heavy computation for DCR. In stages 2-4, our PWRF models modal-shared and modal-specific details of different auxiliary modalities, which are further integrated with the primary RGB modality. The outputs of four stages are fed to Segformer head \cite{xie2021segformer} for semantic segmentation.}
	\label{fig:network}
\end{figure*}

To explore the contributions of our PWRF for SMM semantic segmentation, which targets at segment semantic objects under the synthetic RGB-depth-event-LiDAR condition, we plug our PWRF framework in the baseline \cite{zhang2023delivering}. As shown in Fig. \ref{fig:network}, the proposed PWRF based synthetic multi-modal  semantic segmentation network consists of four stages. In each stage, RGB and the remaining modalities are taken as the primary and auxiliary modalities, respectively. For the primary RGB modality, the Multi-Head Self-Attention (MHSA) \cite{xie2021segformer} is used to extract deep features. For the auxiliary modalities, their fusion is solved by the proposed PWRF framework, on top of which modal-shared and modal-specific details are further integrated via the designed shared-specific interaction module. In the final, the primary RGB modality and the integrated auxiliary modalities details are fed in the segmentation head \cite{xie2021segformer} for semantic segmentation. On top of the modal-shared and modal-specific details from PWRF, the shared-specific integration module will be illustrated in the following. In Fig. \ref{fig:network}, feature refine fusion module, parallel pooling mixer blocks, and token-select hub can refer to \cite{zhang2023delivering}, which are not detailed due to they being not our contributions.



\subsubsection{Shared-specific Integration}

In view of the complementary characteristic of modal-shared and modal-specific cues with respect to the input, their primitive integration will benefit the semantic exploration. To this end, we design a shared-specific integration module to combine complementaries of modal-shared and modal-specific details for better semantic inference, which consists of two components, including primitive modal-specific details generation and shared-specific interaction. 

\textbf{A. Primitive modal-specific details generation}

{In order to attenuate noise of each modal-specific information, primitive modal-specific information is generated by using the original modality features ${\bf{f}}_i^n$ and modal-shared details $\mathbf{F}_{i,{shd}}$ as follows}
\begin{equation}
	\label{equ:spccap-cat}
	{\hat{\bf{F}}_{i,{sp}}^n \in {\Re ^{{H_i} \times {W_i} \times C}}= ConvCate\left( {\left\{ {{\bf{f}}_i^n,{\bf{SP}}_i^n} \right\},dim  = 3} \right),}
\end{equation}
\begin{equation}
	\label{equ:spc-cap-fea}
	{\mathbf{F}_{i,{sp}}^n \in {\Re ^{{H_i} \times {W_i} \times C}} = ConvCate\left( {\left\{ {\mathbf{F}_{i,{shd}},{\bf{SP}}_i^n} \right\},dim  = 3} \right),}
\end{equation}
{where the modal-shared details, denoted as \(\mathbf{F}_{i, {shd}}\), are derived from reshaping the whole-level modal capsules $\mathbf{WP}_{i}$. 
}
The primitive modal-shared details can be derived as 
\begin{equation}
	\label{equ:spc-cap}
	{\mathbf{F}_{i,{psg}}^n \in {\Re ^{{H_i} \times {W_i} \times C}} = \sigma \left( \hat{\bf{F}}_{i,{sp}}^n \right) \odot \mathbf{F}_{i,{sp}}^n + \mathbf{F}_{i,{sp}}^n.}
\end{equation}

As such, a more primitive modal-specific information is achieved by combining three single modalities together as follows
\begin{equation}
	\label{equ:spc-capall}
	{\mathbf{F}_{i,{psg}} \in {\Re ^{{H_i} \times {W_i} \times C}} = ConvCate\left( {{{\left. \mathbf{F}_{i,{psg}}^n \right|}_{n = 1,2,3}},dim  = 3} \right).}
\end{equation}


\textbf{B. Shared-specific interaction}

{To integrate modal-shared and modal-specific details, we propose a shared-specific interaction module. Besides modal-shared details $\mathbf{F}_{i,{shd}}$ and primitive modal-specific details $\mathbf{F}_{i,{psg}}$, we also employ the selected modal details of Self-Query Hub\cite{zhang2023delivering} denoted as $\mathbf{F}_{i,{sqh}}$.} Specifically, three parallel branches are first designed to interact these three components. Within each branch, one component is selected as the primary cue, while the remaining two components are utilized to attend the primary cue for better semantic exploration, which is achieved by a spatial attention and a channel attention. The spatial attention is implemented as follows
\begin{equation}
	\label{equ: sa}
	{\bf{SA}}_i = \sigma \left( {Conv(CGMP\left( {{\bf{CP}}_i^1 + {\bf{CP}}_i^2 + {\bf{CP}}_i^3} \right), dim=3)} \right),
\end{equation}
where ${\bf{CP}}^1_i$ is the primary component. ${\bf{CP}}^2_i$ and ${\bf{CP}}^3_i$ are the remaining two components. {$CGMP(*)$ denotes the global max pooling operation performed along the channel direction.}

The channel attention is implemented as 
\begin{equation}
	\label{equ: ca}
	{\bf{CA}}_i = \sigma \left( {Conv(GMP\left( {{\bf{CP}}_i^1 \odot {\bf{SA}}_i} + {\bf{CP}}^1_i \right), dim=3)} \right) ,
\end{equation}
{where $GMP(*)$ refers to the adaptive global max pooling operation.}

Based on the spatial and channel attentions, the primary component can be attended as
\begin{equation}
	\label{equ: pa}
	{{\bf{F}}_{i,ssi}^1 = {\bf{CP}}_i^1 \odot {\bf{CA}}_i + {\bf{CP}}_i^1.}
\end{equation}

Doing Eqs.\ (\ref{equ: sa})-(\ref{equ: pa}), we obtain the shared-specific interaction for each primary component ${\bf{CP}}_i^j,j=1,2,3$. 


{The interacted component ${\bf{F}}_{i,ssi}$ in three branches are integrated to get the merged information using the element-wise multiplication and element-wise addition, which is written as}
\begin{equation}
	\label{mu}
	{{\bf{F}}_i^{u} = ConvCate\left( {\left\{ {{{\left. { \otimes \left\{ {\bf{F}}_{i,ssi}^j \right\}} \right|}_{j = 1,2,3}},{{\left. { \oplus \left\{ {\bf{F}}_{i,ssi}^j \right\}} \right|}_{j = 1,2,3}}} \right\}} \right),}
\end{equation}
where $ \oplus $ represents the element-wise addition. 
\subsubsection{Model training}
{On top of the fusion features $	{\bf{F}}_i^{u}$, we integrate it with the primary RGB modality using the fusion step \cite{zhang2023delivering} to get the multi-modal fusion features, which is fed into the multi-layer perception decoder \cite{xie2021segformer} to predict the semantic result ${\bf{Pre}}$.} To train the model, the online hard example mining cross-entropy loss function \cite{shrivastava2016training} is used to compute the difference between the semantic predictions and the ground truth ${\bf{GT}}$, \emph{i.e.},
\begin{equation}
	Loss = OHEMCrossEntropyLoss({\bf{Pre}}, {\bf{GT}}).
\end{equation}
\subsection{VDT salient object detection}
To explore the effectiveness of our PWRF for VDT salient object detection, we design a network as shown in Fig. \ref{fig:VDT}. To be concrete, Swin-Transformer \cite{liu2021swin} is utilized to learn the backbone features of triple modalities, which are further fed in our PWRF to get modal-shared and modal-specific semantics. After that, a stacking adjacent-scale attention decoder is designed to integrate different-scale modal-shared/specific semantics. The predictions of these two sub-decoders are combined to achieve the final saliency map. The following will detail the Adjacent-Scale Attention (ASA) and stacking ASA decoder.
\begin{figure}[htbp]
		\centering
		\includegraphics[width=0.92\linewidth]{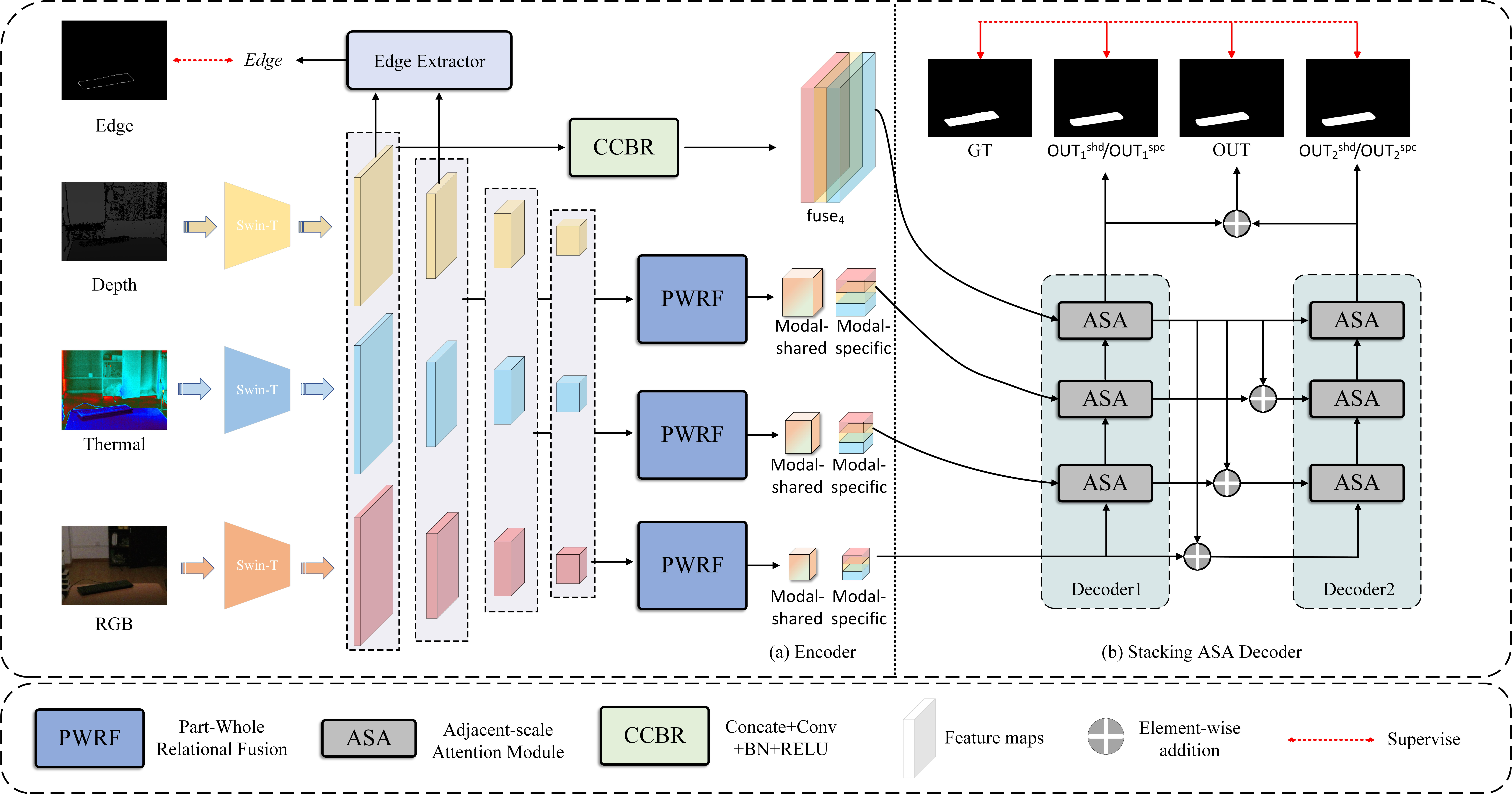}
		\caption{VDT salient object detection framework based on PWRF. Swin-Transformer \cite{liu2021swin} is utilized to learn the backbone features of triple modalities, which are further fed in our PWRF to get modal-shared and modal-specific semantics. After that, a stacking adjacent-scale attention decoder is designed to integrate different-scale modal-shared/specific semantics. The predictions of these two sub-decoders are combined to achieve the final saliency map.}
		\label{fig:VDT}
\end{figure}
\subsubsection{Adjacent-scale attention}
High-level features contain rich semantic information, encapsulating the overall properties of salient objects. In contrast, low-level features preserve edge details of salient objects. To complement their superiority, we design an ASA module to integrate adjacently high-level and low-level semantics for modal-shared and modal-specific in Fig. \ref{fig:mam}, which is composed by three components, including adjacent-scale integration, dual-branch attention, and selective aggregation.
\begin{figure*}[htbp]
	\centering
	\includegraphics[width=0.86\linewidth]{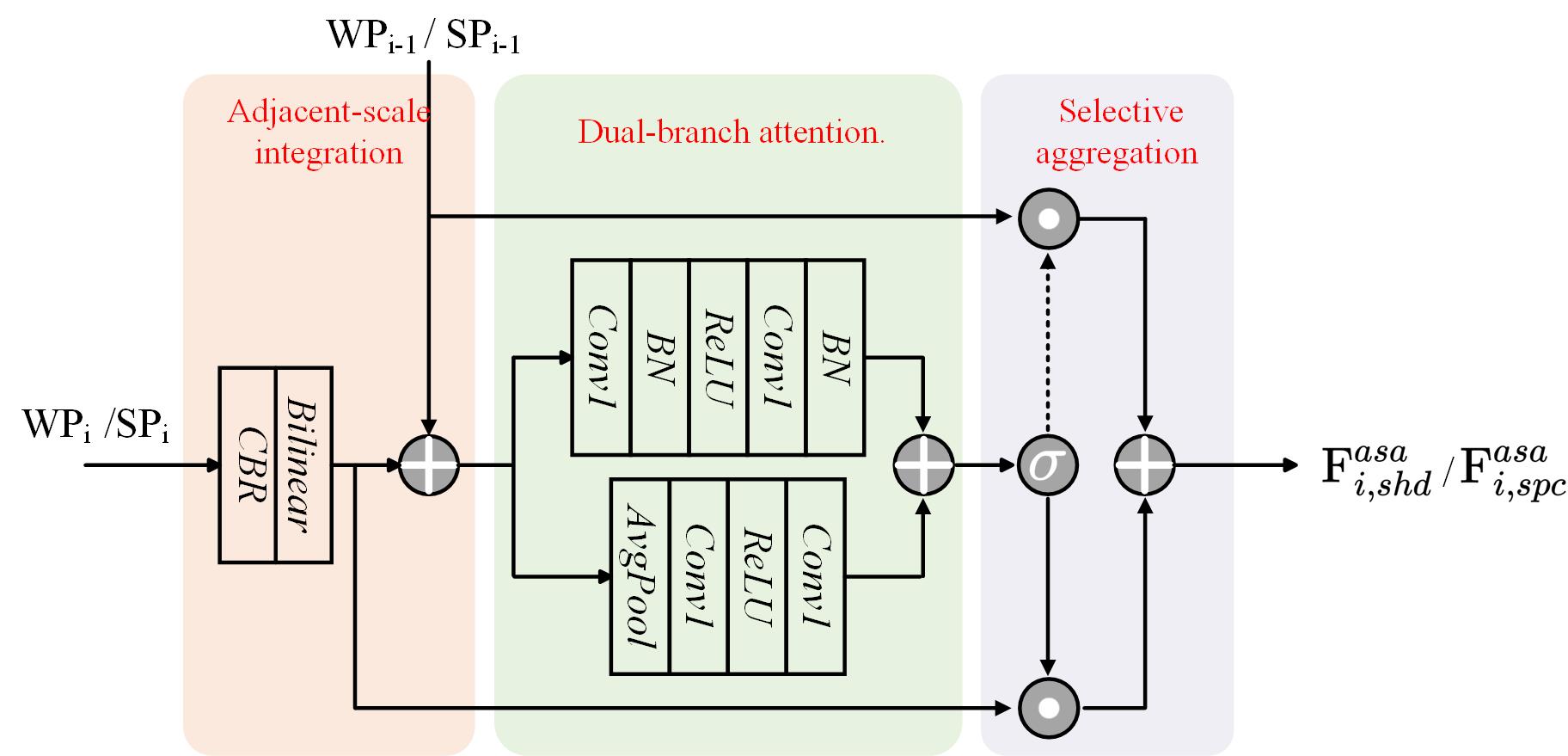}
	\caption{Adjacent-scale Attention Module, which is composed by three components, including adjacent-scale integration, dual-branch attention, and selective aggregation.}
	\label{fig:mam}
\end{figure*}

\textbf{Adjacent-scale integration.} {The adjacent-level modal-shared details (${\bf{WP}}_{i-1}$ and ${\bf{WP}}_i$) and modal-specific details (${{\bf{SP}}_{i-1}}$ and ${{\bf{SP}}_i}$) are integrated via }
\begin{equation}
	{{{\bf{F}}_{{i-1, shd}}} = {\bf{WP}}_{i-1} \oplus {\bf{WP}}_i^\prime.}
\end{equation}
\begin{equation}
	{{{\bf{F}}_{{i-1}, spc}} = {{\bf{SP}}_{i-1}} \oplus {{\bf{SP}}_i}^\prime.}
\end{equation}

{${\bf{WP}}_i^\prime$ and ${\bf{SP}}_i^\prime$ can be obtained by}
\begin{equation}
	{\bf{WP}}_i^\prime  = Bilinear\left( {CBR\left( {\bf{WP}}_i \right)} \right),
\end{equation}
\begin{equation}
	{\bf{SP}}_i^\prime  = Bilinear\left( {CBR\left( {{\bf{SP}}_i} \right)} \right),
\end{equation}
where $CBR(\cdot)$ and $Bilinear$ represent the operations of (Convolution + BatchNorm + ReLU) and bilinear upsampling, respectively.

\textbf{Dual-branch attention.} A dual-branch attention including local attention and global attention is designed to attend the informative regions. Specifically, the dual-branch attention is achieved by
\begin{equation}
	{{{\bf{F}}^{dba}_{i,shd}} = CBRCB\left( {{\bf{F}}_{{i, shd}}} \right) + ACRC\left( {{\bf{F}}_{{i, shd}}} \right),}
\end{equation}
\begin{equation}
	{{{\bf{F}}^{dba}_{i,spc}} = CBRCB\left( {{\bf{F}}_{{i, spc}}} \right) + ACRC\left( {{\bf{F}}_{{i, spc}}} \right),}
\end{equation}
where $CBRCB (\cdot)$ is the local attention using (Convolution + BatchNorm + ReLU + Convolution + BatchNorm). $ACRC (\cdot)$ is the global attention using (Average pooling + Convolution + ReLU + Convolution).

\textbf{Selective aggregation.} To address the feature discrepancy, a selective aggregation strategy is designed to suppress redundant information and prevent feature contamination. To this end, a gate signal mechanism is introduced to aggregate adjacent-level modal-shared and modal-specific details, \emph{i.e.},
\begin{equation}
	{{{\bf{F}}^{asa}_{i,shd}} = {\bf{WP}}_{i}^\prime \otimes \sigma \left( {{{\bf{F}}^{dba}_{i,{shd}}}} \right) + {\bf{WP}}_{i-1} \otimes \left( 1 - \sigma \left( {{\bf{F}}^{dba}_{i,{shd}}} \right) \right),}
\end{equation}
\begin{equation}
	{{{\bf{F}}^{asa}_{i,spc}} =  {\bf{SP}}_{i}^\prime \otimes \sigma \left( {{\bf{F}}^{dba}_{i,spc}} \right) + {\bf{SP}}_{i-1} \otimes \left( {1 - \sigma \left( {{\bf{F}}^{dba}_{i,spc}} \right)} \right).}
\end{equation}

\subsubsection{Stacking ASA decoder}

As illustrated in Fig. \ref{fig:VDT}(b), we stack two sub-decoders composed by ASA to improve the feature aggregation to produce saliency maps, which implement features aggregation following bottom-up and top-down flows. Specifically in the bottom-up process, the ASA within each decoder progressively aggregates from high-level to low-level features for both modal-shared and modal-specific, separately. {The resulting aggregated features ${{\bf{F}}^{asa}_{i,shd}}$ and ${{\bf{F}}^{asa}_{i,spc}}$ contribute to the generation of a preliminary saliency map.} Conversely, in the top-down process, the shallowest aggregated features of the first sub-decoder are densely used to guide the second sub-decoder to learn primitive features. By the way, edge cues are employed to enhance the depth feature maps by adjusting their poor conditions. This process improves boundary refinement, allowing for more accurate delineation of object edges and a better representation of spatial structures within the depth maps. Aggregated features of two sub-decoders are combined to generate the final saliency prediction.

\subsubsection{Model training}
The loss function $L$ is defined as: 
\begin{equation}
	L =\sum\limits_{i = 1}^5 \left(  L_B(\mathbf{Pre}_i,\mathbf{GT}) + L_S(\mathbf{Pre}_i,\mathbf{GT}) + L_I(\mathbf{Pre}_i,\mathbf{GT})\right), 
\end{equation} 
where $L_B$, $L_S$, and $L_I$ represent binary cross entropy loss \cite{de2005tutorial}, structural similarity loss \cite{li2021salient}, and intersection-over-union loss \cite{rahman2016optimizing}, respectively. $\mathbf{Pre}_i$ represent the $ith$ predicted saliency map. $\mathbf{GT}$ is the Ground Truth. 
\subsection{Intermediate Visualization of PWRF}
To get a more intuitive perception for the effectiveness of our PWRF, as shown in Fig. \ref{fig:visual-fea-map}, modal-shared and primitive modal-specific features are visualized. In Fig. \ref{fig:visual-fea-map}, modal-shared knowledge can well capture the common details of three modalities while with blurry boundaries. In contrast, modal-specific features complement to focus on object shapes with clear boundaries. It is obvious for the necessity of integrating modal-shared and modal-specific details for further decision.
\begin{figure*}[htbp]
	\centering
	\includegraphics[width=1\linewidth]{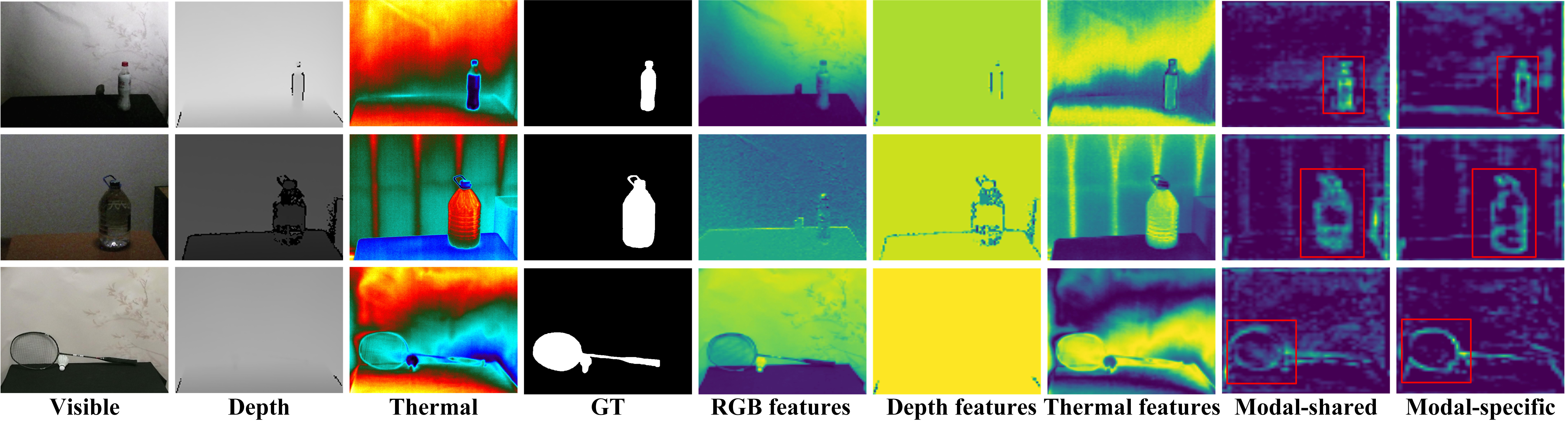}
	\caption{Visualization of modal-shared and primitive modal-specific features.  Modal-shared knowledge well capture the common details of three modalities. Modal-specific features focus on different cues such as object shape.}
	\label{fig:visual-fea-map}
\end{figure*}
\section{Experiment and Analysis}
\label{sec:Experiment}
In this section, we will discuss the experimental results of the proposed methods for the tasks of SMM semantic segmentation and VDT salient object detection.
\subsection{Datasets}
\textbf{DELIVER \cite{zhang2023delivering}} is a large-scale dataset for synthetic multi-modal semantic segmentation, including RGB, Depth, LiDAR, and Event, which was created using the CARLA simulator. Each image resolution is 1042\texttimes1042. It contains 47,310 frames, with 7,885 front-view samples divided to 3,983/2,005/1,897 samples for training/validation/testing, respectively. The dataset introduces adverse conditions and sensor failure scenarios, such as environmental variations and partial sensor malfunctions, which is a challenge for autonomous driving. 

\textbf{VDT-2048 dataset \cite{song2022novel}} consists of 2048 images with pixel-wise annotations (ground truths) or VDT salient object detection. The dataset is divided into 1048 training images and 1000 testing images.

\subsection{Implementation details}
\textbf{Multi-modal synthetic semantic segmentation.} We conduct our SMM semantic segmentation model trained on four A100 GPUs with an initial learning rate (LR) set to 6e-5, using the poly strategy with a power of 0.9. The LR is adjusted to 0.1$\times$ the original LR for the first 10 epochs for warming up. AdamW optimizer is chosen for training with epsilon set to 1e-8 and weight decay set to 1e-2. Data augmentation techniques include random resizing with a ratio ranging from 0.5 to 2.0, random horizontal flipping, random color jittering, random Gaussian blurring, and random cropping. 

\textbf{VDT salient object detection.} We conduct our VDT salient object detection model trained on an RTX 3090 GPU. During the training process, we resize all training images to $384 \times 384$ and apply data augmentation techniques such as random flipping and clipping. The backbone network parameters are initialized with pre-trained weights from the Swin-B network \cite{liu2021swinnet}. Adam optimizer is chosen to train the model using a batchsize of 4 and an initial learning rate of 5e-5. The learning rate is decreased by a factor of 10 every 80 epochs.
\subsection{Evaluation Metrics}
To quantitatively evaluate the performance on DELIVER \cite{zhang2023delivering}, we select mIoU as the metric, which is in accordance with the state-of-the-arts.

To quantitatively evaluate different saliency models on the VDT-2048 dataset, we use 10 comprehensive evaluation metrics including S-measure (S) \cite{fan2017structure}, Mean Absolute Error (MAE) \cite{achanta2009frequency}, various F-measure metrics \cite{achanta2009frequency} ($F_{\beta}^{mean}$, $F_{\beta}^{adp}$), and E-measure metrics \cite{fan2018enhanced} ($E_{\xi}^{mean}$, $E_{\xi}^{adp}$). The mathematical definitions of all metrics are depicted as follows,

\textbf{MAE:}
\begin{equation}
	\mathrm{MAE}=\frac{1}{W \times H} \sum_{x=1}^{W} \sum_{y=1}^{H}|\mathbf{P}(x, y)-\mathbf{G}(x, y)|,
\end{equation}
where $\mathbf{P}(x, y) $ and $\mathbf{G}(x, y)$ predicted saliency map and ground truth respectively.
  
\textbf{F-measure:}



\begin{equation}
	F_{\beta} =  (1 + \beta^2) \cdot \frac{Precision \cdot Recall}{\beta^2 \cdot Precision + Recall},
\end{equation}
where $\beta^2$ is the parameter used to balance $Precision$ and $Recall$.

\textbf{E-measure:}
%
%
%
\begin{equation}
	E_{\xi} = \frac{1}{W \times H} \sum_{x=1}^{W} \sum_{y=1}^{H} \theta(x, y),
\end{equation}
where $\theta$ represents the relation between predicted map and ground truth. $H \times W$ is the spatial resolution of the input.

\textbf{S-measure:}
\begin{equation}
	\mathrm{S} = \alpha S_o - (1 - \alpha) S_r,
\end{equation}
where $S_o$ and $S_r$ denote object-aware and region-aware structural similarity, respectively. $\alpha$ is set to 0.5.
\subsection{Comparison against the State of the Art}
\subsubsection{SMM semantic segmentation}
Table \ref{tab:cmp-del} provides a comprehensive comparison between our approach and some State-Of-The-Art (SOTA) methods, including HRFuser \cite{broedermann2023hrfuser}, SegFormer \cite{xie2021segformer}, TokenFusion \cite{wang2022multimodal}, CMX \cite{zhang2023cmx}, and CMNeXt \cite{zhang2023delivering}, on the DELIVER dataset \cite{zhang2023delivering}. Following \cite{zhang2023delivering}, we list evaluation values of different models on the validation set of DELIVER \cite{zhang2023delivering} in Table \ref{tab:cmp-del}. In Table \ref{tab:cmp-del}, most cross-modal fusion approaches are beaten by the multi-modal fusion methods. Besides, our model achieves the best IoU values over the multi-modal fusion methods, which proves the superiority of our model over the other methods. To be more concrete, the IoU value for each class are listed in Table \ref{tab: classdel}. For more comprehensive comparison, we test our model and the SOTA multi-modal fusion method CMNeXt \cite{zhang2023delivering} on the test set of the DELIVER dataset \cite{zhang2023delivering}. The mIoU values of our model and CMNeXt \cite{zhang2023delivering} are 54.29\% and 53\%, respectively, which proves a significant improvement of our model over the SOTA method.
\begin{table}[htbt]
	\centering
	\caption{mIoU values of different models on DeLIVER dataset \cite{zhang2023delivering}.}
	\label{tab:cmp-del}
	\begin{tabular}{@{}l|l|l|c@{}}
		\toprule
		Method        & Modal        & Backbone      & DeLIVER                                                \\ \midrule
		HRFuser \cite{broedermann2023hrfuser}   & RGB-D    & HRFormer-T \cite{broedermann2023hrfuser}    & 51.88                                                  \\
		TokenFusion \cite{wang2022multimodal} & RGB-D    & MiT-B2 \cite{xie2021segformer}       & 60.25                                                  \\
		CMX \cite{zhang2023cmx}      & RGB-D    & MiT-B2        & 62.67                                                  \\
		CMNeXt \cite{zhang2023delivering}        & RGB-D    & MiT-B2 \cite{xie2021segformer}       & 63.58                                                  \\
		HRFuser \cite{broedermann2023hrfuser}   & RGB-E    & HRFormer-T \cite{broedermann2023hrfuser}   & 42.22                                                  \\
		TokenFusion \cite{wang2022multimodal} & RGB-E    & MiT-B2 \cite{xie2021segformer}       & 45.63                                                  \\
		CMX \cite{zhang2023cmx}      & RGB-E    & MiT-B2 \cite{xie2021segformer}       & 56.52                                                  \\
		CMNeXt \cite{zhang2023delivering}        & RGB-E    & MiT-B2 \cite{xie2021segformer}       & 57.48                                                  \\
		HRFuser \cite{broedermann2023hrfuser}   & RGB-Li    & HRFormer-T \cite{broedermann2023hrfuser}   & 43.13                                                  \\
		TokenFusion \cite{wang2022multimodal} & RGB-Li    & MiT-B2 \cite{xie2021segformer}       & 53.01                                                  \\
		CMX \cite{zhang2023cmx}      & RGB-Li    & MiT-B2        & 56.37                                                  \\
		CMNeXt \cite{zhang2023delivering}        & RGB-Li    & MiT-B2 \cite{xie2021segformer}       & 58.04                                                  \\
		HRFuser \cite{broedermann2023hrfuser}   & RGB-D-E-Li   & HRFormer-T \cite{broedermann2023hrfuser}   & 52.97                                                  \\
		CMNeXt \cite{zhang2023delivering}        & RGB-D-E-Li   & MiT-B2 \cite{xie2021segformer}       & 66.30                                                  \\    \hline
		\addlinespace
		OURS        & RGB-D-E-Li   & MiT-B2 \cite{xie2021segformer}           &  \bf{ 66.47}                                                  \\ \bottomrule
	\end{tabular}
\end{table}

\begin{table}[ht]
	\centering
	\caption{IoU, F1, and Accuracy for Different Classes in the DeLIVER dataset \cite{zhang2023delivering}.}
	\label{tab: classdel}
	\label{metric for classes}
	\begin{tabular}{@{}l|c|c|c@{}}
		\toprule
		\textbf{Class}   & \textbf{IoU (\%)} & \textbf{F1 (\%)} & \textbf{Accuracy (\%)} \\ \midrule
		Building         & 89.28             & 94.34           & 98.29                  \\
		Fence            & 44.4              & 61.5            & 59.45                  \\
		Other            & 0                 & 0               & 0                      \\
		Pedestrian       & 75.94             & 86.33           & 85.78                  \\
		Pole             & 75.61             & 86.11           & 85.2                   \\
		RoadLine         & 86.17             & 92.57           & 90.63                  \\
		Road             & 98.33             & 99.16           & 98.94                  \\
		SideWalk         & 80.87             & 89.43           & 96.11                  \\
		Vegetation       & 89.17             & 94.28           & 93.92                  \\
		Cars             & 88.8              & 94.07           & 98.57                  \\
		Wall             & 64.45             & 78.38           & 88.45                  \\
		TrafficSign      & 72.4              & 83.99           & 77.28                  \\
		Sky              & 99.43             & 99.71           & 99.75                  \\
		Ground           & 2.77              & 5.38            & 4.2                    \\
		Bridge           & 51.22             & 67.74           & 59.4                   \\
		RailTrack        & 54.38             & 70.45           & 73.97                  \\
		GroundRail       & 48.82             & 65.61           & 50.14                  \\
		TrafficLight     & 83.19             & 90.82           & 87.28                  \\
		Static           & 33.04             & 49.67           & 34.98                  \\
		Dynamic          & 34.29             & 51.07           & 49.79                  \\
		Water            & 42.11             & 59.27           & 42.4                   \\
		Terrain          & 84.52             & 91.61           & 93.79                  \\
		TwoWheeler       & 75.83             & 86.25           & 87.12                  \\
		Bus              & 92.69             & 96.21           & 95.95                  \\
		Truck            & 93.97             & 96.89           & 96.89                  \\ \hline
		\addlinespace
		Mean             & 66.47             & 75.63           & 73.93                  \\ \bottomrule
	\end{tabular}
\end{table}
\begin{table*}[htbp]
	\centering
	\caption{Results on adverse conditions in the DeLIVER dataset \cite{zhang2023delivering}. Sensor failure cases are MB: Motion Blur; OE: Over-Exposure; UE: Under-Exposure; LJ: LiDAR-Jitter; and EL: Event Low-resolution.}
	\label{tab:adverse_conditions}
	\scalebox{0.68}{\begin{tabular}{l|ccccc|ccccc|c}
			\toprule
			Model-modality & Cloudy & Foggy & Night & Rainy & Sunny & MB & OE & UE & LJ & EL & Mean \\
			\midrule
			HRFuser-RGB \cite{broedermann2023hrfuser} & 49.26 & 48.64 & 42.57 & 50.61 & 50.47 & 48.33 & 35.13 & 26.86 & 49.06 & 49.88 & 47.95 \\
			SegFormer-RGB \cite{xie2021segformer} & 59.99 & 57.30 & 50.45 & 58.69 & 60.21 & 57.28 & 56.64 & 37.44 & 57.17 & 59.12 & 57.20 \\
			TokenFusion-RGB-D \cite{wang2022multimodal} & 50.92 & 52.02 & 43.37 & 50.70 & 52.21 & 49.22 & 46.22 & 36.39 & 49.58 & 49.17 & 49.86 \\
			CMX-RGB-D \cite{zhang2023delivering} & 63.70 & 62.77 & 60.74 & 62.37 & 63.14 & 59.50 & 60.14 & 55.84 & 62.65 & 63.26 & 62.66 \\
			HRFuser-RGB-D-E-L \cite{broedermann2023hrfuser} & 56.20 & 52.39 & 49.85 & 52.53 & 54.02 & 49.44 & 46.31 & 46.92 & 53.94 & 52.72 & 52.97 \\
			CMNeXt-RGB-D-E-L \cite{zhang2023delivering} & 68.70 & \bf{65.67} & 62.46 & \bf{67.50} & 66.57 & 62.91 & 64.59 & 60.00 & 65.92 & 65.48 & 66.30 \\ 
			PWRF-RGB-D-E-L & \bf{69.53} & 65.11 & \bf{64.05} & 65.8 & \bf{67.5} & \bf{63.02} & \bf{64.84} & \bf{60.37} & \bf{66.2} & \bf{67.14} &  \bf{66.47} \\ 
			\bottomrule
	\end{tabular}}
\end{table*}
\begin{table*}[htbp]
	\centering
	\caption{Quantitative comparison results (\%) of $S$, $F_{\beta}^{mean}$, $F_{\beta}^{adp}$, $E_{\xi}^{mean}$, $E_{\xi}^{adp}$, and $MAE$ on the VDT-2048 dataset. Here, ``$\uparrow$'' (``$\downarrow$'') means that the larger (smaller) the better. The best three results in each column are marked in \textcolor{red}{red}, \textcolor{green}{green}, and \textcolor{blue}{blue}, respectively. {\textbf{Note}: \textcolor{red}{Red} indicates the best performance in each metric.}}
	\label{cmp-vdt}
	\resizebox{\textwidth}{!}{%
		\begin{tabular}{l|c|cccccc}
			\toprule
			\textbf{Methods} & \textbf{Type} &  $S\uparrow$ & $MAE\downarrow$ & \textbf{$E_{\xi}^{adp} \uparrow$} & \textbf{$E_{\xi}^{mean} \uparrow$} & \textbf{$F_{\beta}^{adp} \uparrow$} & \textbf{$F_{\beta}^{mean} \uparrow$} \\
			\midrule
			CPD \cite{wu2019cascaded} & V & 90.44 & 0.39 & 92.70 & 95.01 & 76.45 & 83.76 \\
			RAS \cite{chen2018reverse} & V & 89.00 & 0.40 & 96.15 & 96.50 & 80.79 & 82.92 \\
			\midrule
			BBSNet \cite{fan2020bbs} & VD & 91.17 & 0.46 & 87.47 & 93.57 & 69.57 & 82.67 \\
			DPANet \cite{chen2020dpanet} & VD & 72.26 & 1.92 & 53.28 & 72.22 & 29.19 & 48.52 \\
			RD3D \cite{chen20223} & VD & 90.95 & 0.47 & 83.54 & 92.31 & 64.62 & 81.20 \\
			SwinNet \cite{liu2021swinnet} & VD & 91.98 & 0.37 & 89.78 & 95.07 & 73.21 & 84.58 \\
			HRTransNet \cite{tang2022hrtransnet} & VD & 91.44 & 0.31 & 96.17 & 97.60 & \textcolor{blue}{88.27} & 85.49 \\
			\rowcolor[gray]{0.9}
			Ours     & VD    & 90.84 &  0.33  &  97.57  &  97.61   &  84.77  &   85.51   \\
			\midrule
			CGFNet \cite{wang2021cgfnet} & VT & 91.66 & 0.33 & 93.19 & 94.47 & 78.22 & 84.80 \\
			CSRNet \cite{li2021salient} & VT & 88.21 & 0.50 & 94.94 & 95.57 & 78.88 & 82.78 \\
			DCNet \cite{tu2022weakly} & VT & 87.87 & 0.38 & 96.58 & 94.36 & 85.21 & 84.5 \\
			LSNet \cite{zhou2023lsnet} & VT & 88.67 & 0.44 & 93.27 & 96.31 & 76.07 & 80.97 \\
			SwinNet\cite{liu2021swinnet} & VT & \textcolor{green}{93.70} & \textcolor{blue}{0.26} & 94.44 & 97.46 & 80.90 & 88.87 \\
			HRTransNet\cite{tang2022hrtransnet} & VT & 92.81 & \textcolor{blue}{0.26} & 96.80 & 98.09 & 84.46 & 87.59 \\
			\rowcolor[gray]{0.9}
			Ours     & VT    & 92.85 &  \textcolor{blue}{0.26}  &  \textcolor{green}{98.65}  &  \textcolor{blue}{98.43}   &  \textcolor{green}{89.02}  &  89.45   \\
			\midrule
			HWSI \cite{song2022novel} & VDT & 93.18 & \textcolor{blue}{0.26} & 98.15 & \textcolor{green}{98.45} & 87.18 & \textcolor{blue}{89.61} \\
			MFFNet \cite{wan2023mffnet} & VDT & \textcolor{red}{93.94} & \textcolor{green}{0.25} & \textcolor{blue}{98.31} & 98.25 & 87.58 & \textcolor{green}{90.34} \\
			\midrule
			\midrule
			\rowcolor[gray]{0.9}
			Ours & VDT & \textcolor{blue}{93.27} & \textcolor{red}{0.23} & \textcolor{red}{98.84} & \textcolor{red}{98.52} & \textcolor{red}{90.17} & \textcolor{red}{90.38} \\
			\bottomrule
		\end{tabular}%
	}
\end{table*}

\begin{table*}[htbp]
	\centering
	\caption{Quantitative results (\%) in V challenges. Here, ``$\uparrow$'' (``$\downarrow$'') means that the larger (smaller) the better. The best three results in each column are marked in \textcolor{red}{red}, \textcolor{green}{green}, and \textcolor{blue}{blue}, respectively. {\textbf{Note}: \textcolor{red}{Red} indicates the best performance in each metric.}}
	\label{cmp-V}
	\Huge
	\resizebox{\textwidth}{!}{%
		\Huge 
		\begin{tabular}{l|cccc|cccc|cccc|cccc|cccc|cccc|cccc}  
			\toprule
			\multirow{3}{*}{Method} & \multicolumn{4}{c}{\textbf{V-BSO}} & \multicolumn{4}{|c}{\textbf{V-LI}} & \multicolumn{4}{|c}{\textbf{V-MSO}} & \multicolumn{4}{|c}{\textbf{V-NI}} & \multicolumn{4}{|c}{\textbf{V-SA}} & \multicolumn{4}{|c}{\textbf{V-SI}} & \multicolumn{4}{|c}{\textbf{V-SSO}} \\
			\cmidrule(lr){2-5} \cmidrule(lr){6-9} \cmidrule(lr){10-13} \cmidrule(lr){14-17} \cmidrule(lr){18-21} \cmidrule(lr){22-25} \cmidrule(lr){26-29}
			&  $S\uparrow$ & $MAE\downarrow$ & $E_{\xi}^{adp}\uparrow$  & \textbf{$F_{\beta}^{adp} \uparrow$}  &  $S\uparrow$ & $MAE\downarrow$ & $E_{\xi}^{adp}\uparrow$  & \textbf{$F_{\beta}^{adp} \uparrow$}  &  $S\uparrow$ & $MAE\downarrow$ & $E_{\xi}^{adp}\uparrow$  & \textbf{$F_{\beta}^{adp} \uparrow$}  &  $S\uparrow$ & $MAE\downarrow$ & $E_{\xi}^{adp}\uparrow$  & \textbf{$F_{\beta}^{adp} \uparrow$}  &  $S\uparrow$ & $MAE\downarrow$ & $E_{\xi}^{adp}\uparrow$  & \textbf{$F_{\beta}^{adp} \uparrow$}  &  $S\uparrow$ & $MAE\downarrow$ & $E_{\xi}^{adp}\uparrow$  & \textbf{$F_{\beta}^{adp} \uparrow$}  &  $S\uparrow$ & $MAE\downarrow$ & $E_{\xi}^{adp}\uparrow$  & \textbf{$F_{\beta}^{adp} \uparrow$}   \\
			\midrule
			BBSNet \cite{fan2020bbs}     & 95.98 & 0.77 & 98.94  & 92.64  & 89.87 & 0.64 & 86.46  & 67.20   & 91.31 & 0.53 & 91.59  & 75.30   & 82.54 & 0.80  & 75.57  & 52.34   & 92.13 & 0.34 & 90.67  & 72.06  & 89.29 & 0.65 & 86.29  & 66.80  & 85.48 & 0.28 & 65.83 & 41.83  \\
			CGFNet \cite{wang2021cgfnet} & 95.96 & 0.71 & 99.19  & 93.77  & 90.35 & 0.46 & 93.63  & 77.55  & 91.18 & 0.49 & 94.77  & 80.20   & 88.40 & 0.35  & 86.93   & 68.57  & 92.10 & 0.30 & 92.90 & 76.67   & 91.19 & 0.43 & 93.53  & 77.84  & 84.21 & 0.12 & 79.42  & 56.03  \\
			CPD \cite{wu2019cascaded}    & 94.29 & 0.92 & 98.58  & 93.18 & 88.92 & 0.55 & 93.07  & 75.50   & 90.65 & 0.46 & 95.59  & 80.12   & 81.11 & 0.55  & 84.02  & 59.55   & 91.96 & 0.30 & 94.65  & 79.65   & 89.29 & 0.55 & 92.95 & 74.29  & 84.78 & 0.12 & 77.80  & 53.16  \\
			CSRNet \cite{li2021salient}  & 90.94 & 1.39 & 96.15  & 89.85 & 87.73 & 0.58 & 95.72  & 79.34  & 85.98 & 0.89 & 94.14  & 79.40   & 85.89 & 0.43  & 92.29   & 72.60   & 86.15 & 0.47 & 95.11  & 75.69   & 86.71 & 0.78 & 94.15  & 77.77  & 81.89 & 0.14 & 83.10  & 56.74  \\
			DCNet \cite{tu2022weakly}    & 94.52 & 0.82 & 99.06  & 94.44 & 87.10 & 0.48 & 97.04 & 84.13   & 88.11 & 0.54 & 98.00 & 85.11   & 82.42 & 0.38  & 92.07  & 78.35   & 89.60 & 0.32 & \color{green}{98.26}  & 83.79  & 88.55 & 0.47 & 98.17  & 85.46  & 77.40 & 0.12 & 92.66 & 72.01    \\
			DPANet \cite{chen2020dpanet} & 77.27 & 4.18 & 93.12 & 69.80 & 71.93 & 2.15 & 54.10  & 29.53  & 71.82 & 2.64 & 58.67  & 31.95   & 63.14 & 2.55  & 44.28   & 20.84   & 71.13 & 1.66 & 45.61  & 21.45   & 71.07 & 2.21 & 52.63  & 27.37 & 62.04 & 1.08 & 30.16  & 5.97   \\
			LSNet \cite{zhou2023lsnet}   & 94.90 & 0.99 & 98.94  & 92.58  & 87.47 & 0.57 & 94.05  & 75.98   & 88.98 & 0.60 & 94.45  & 78.60   & 81.76 & 0.52  & 87.92  & 63.91   & 88.48 & 0.40 & 93.83  & 76.77  & 86.21 & 0.64 & 94.05 & 74.97  & 79.43 & 0.18 & 76.55  & 51.17   \\
			RAS \cite{chen2018reverse}   & 93.88 & 0.95 & 98.24  & 92.36  & 87.23 & 0.57 & 95.61  & 78.08  & 88.67 & 0.55 & 96.78  & 81.98   & 80.71 & 0.52  & 90.60   & 66.15  & 89.33 & 0.33 & 97.11  & 82.12   & 86.40 & 0.65 & 95.86 & 78.07  & 83.07 & 0.13 & 89.18  & 65.19 \\
			RD3D \cite{chen20223}        & 95.52 & 0.91 & 98.60  & 90.81  & 89.46 & 0.62 & 82.77  & 62.15  & 91.31 & 0.57 & 88.83 & 70.76  & 82.63 & 0.67 & 69.80  & 47.31  & 91.48  & 0.38 & 87.22  & 67.05  & 88.66 & 0.70 & 82.42 & 62.10   & 84.66 & 0.24 & 57.07  & 33.14  \\
			SwinNet(VD) \cite{liu2021swinnet} & 96.39 & 0.68 & 99.17  & 93.62  & 90.60 & 0.48 & 90.03  & 72.24 & 91.57 & 0.48 & 92.53  & 76.79  & 84.84 & 0.52 & 81.23  & 59.70 & 93.69 & 0.28 & 92.20  & 75.56 & 89.90 & 0.54 & 90.58 & 72.69   & 86.13  & 0.24 & 66.73  & 42.93  \\
			SwinNet(VT) \cite{liu2021swinnet} & \textcolor{red}{96.76} & \textcolor{green}{0.53} & \textcolor{green}{99.40} & \textcolor{blue}{95.15}  & \textcolor{red}{92.94} & \textcolor{green}{0.35} & 95.03 &  81.19  & \textcolor{green}{92.74} & \textcolor{blue}{0.40} & 95.48  & 82.07  &  \textcolor{green}{90.59} & \textcolor{blue}{0.27} & 91.26   & 74.33  & \textcolor{red}{93.80} & \textcolor{blue}{0.23} & 95.15  & 80.66  &  \textcolor{red}{92.56} &  0.39 & 95.58  & 81.57 & 88.59 &  0.09 & 77.83  & 54.77  \\
			HRTrans(VD)  \cite{tang2022hrtransnet} & 95.72  & {0.64} & 99.26  & 94.63 & 90.43 & {0.42} & 96.42 & 81.57  & 91.24 & {0.42} & 96.91 & 83.39 & {83.98} & {0.42} & 92.16 & 69.74  & 92.40 & {0.25} & 97.37 & 84.25 & 89.81 & {0.49} & 96.52  & 81.45  & 84.63 & {0.12} & 85.45  & 61.91 \\
			HRTrans(VT)  \cite{tang2022hrtransnet}  & 96.15 & \textcolor{blue}{0.57} & 99.24 & 94.75  & 91.84 & 0.37 & 96.94   & 83.75 & 91.70 & \textcolor{green}{0.39} & 97.17  & 84.82  & 89.07 & {0.30} & 94.38  & 77.38 & 92.78 & \textcolor{blue}{0.23} & 97.38  & 84.69  & 91.17 & {0.42} & 96.78  & 83.21   & 85.58 &  {0.11} & 86.93 & 64.92  \\
			HWSI \cite{song2022novel}        & 95.92 & {0.61} & 99.29  & 94.63  & 91.28 & {0.38} & \textcolor{blue}{97.43}  & \textcolor{blue}{84.02} & 92.23 & \textcolor{blue}{0.40} &  \textcolor{blue}{97.60 } &  \textcolor{blue}{86.56} &  \textcolor{blue}{90.51} & {0.28} &  \textcolor{blue}{95.89}  &  \textcolor{blue}{80.28} & 92.48 & {0.24} &  \textcolor{blue}{97.87} &  \textcolor{blue}{85.56} &   \textcolor{blue}{91.86}&  \textcolor{green}{0.38} &  \textcolor{blue}{97.81}  &  \textcolor{blue}{85.08}  &  \textcolor{green}{89.45} &  \textcolor{green}{0.08} &  \textcolor{green}{93.62}  &  \textcolor{green}{75.10} \\
			MFFNet \cite{wan2023mffnet} & \textcolor{green}{96.43}  & \textcolor{blue}{0.57} & \textcolor{blue}{99.37}  & \textcolor{green}{95.28}  & \textcolor{blue}{92.33} & \textcolor{green}{0.35} & \textcolor{green}{98.12}  & \textcolor{green}{86.00} & \textcolor{red}{93.22} & { 0.41} &  \textcolor{green}{98.14}&  \textcolor{green}{87.29}  & \textcolor{red}{91.12} & \textcolor{green}{ 0.26} & \textcolor{red}{97.17} &  \textcolor{green}{82.17}  &  \textcolor{green}{93.76} & \textcolor{green}{ 0.22} &  {97.61}  &  \textcolor{green}{85.88} & \textcolor{green}{92.28} & \textcolor{red}{ 0.37} &  \textcolor{green}{98.17} &  \textcolor{green}{86.07}  & \textcolor{red}{90.70} & \textcolor{red}{0.07} &  \textcolor{blue}{93.51} &  \textcolor{blue}{74.25} \\
			\midrule
			\midrule
			Ours           & \textcolor{blue}{96.19}   & \textcolor{red}{0.51} & \textcolor{red}{99.43}  & \textcolor{red}{95.96}  & \textcolor{green}{92.63} & \textcolor{red}{0.32} & \textcolor{red}{98.86} & \textcolor{red}{88.43} & \textcolor{blue}{92.34} & \textcolor{red}{0.36} & \textcolor{red}{98.91}  & \textcolor{red}{89.32}  & 89.93 & \textcolor{red}{0.25} &  \textcolor{green}{97.13}& \textcolor{red}{84.97} &  \textcolor{blue}{93.66} & \textcolor{red}{0.20} & \textcolor{red}{99.18} & \textcolor{red}{89.24}  & 91.35 & \textcolor{green}{0.38} & \textcolor{red}{98.76}  & \textcolor{red}{87.78} &  \textcolor{blue}{88.97} & \textcolor{red}{0.07} & \textcolor{red}{97.13} & \textcolor{red}{81.34} \\
			\bottomrule
		\end{tabular}%
	}
\end{table*}
\begin{table*}[htbp]
	\centering
	\caption{Quantitative results (\%) in D challenges. Here, ``$\uparrow$'' (``$\downarrow$'') means that the larger (smaller) the better. The best three results in each column are marked in \textcolor{red}{red}, \textcolor{green}{green}, and \textcolor{blue}{blue}, respectively. {\textbf{Note}: \textcolor{red}{Red} indicates the best performance in each metric.}}
	\label{cmp-D}
	\resizebox{\textwidth}{!}{%
		\begin{tabular}{l|cccc|cccc|cccc|cccc}
			\toprule
			\multirow{3}{*}{Method} & \multicolumn{4}{c|}{\textbf{D-BI}} & \multicolumn{4}{c|}{\textbf{D-BM}} & \multicolumn{4}{c|}{\textbf{D-II}} & \multicolumn{4}{c}{\textbf{D-SSO}} \\
			\cmidrule(lr){2-5} \cmidrule(lr){6-9} \cmidrule(lr){10-13} \cmidrule(lr){14-17}
			&  $S\uparrow$ & $MAE\downarrow$ & $E_{\xi}^{adp}\uparrow$  & \textbf{$F_{\beta}^{adp} \uparrow$} &  $S\uparrow$ & $MAE\downarrow$ & $E_{\xi}^{adp}\uparrow$  & \textbf{$F_{\beta}^{adp} \uparrow$} &  $S\uparrow$ & $MAE\downarrow$ & $E_{\xi}^{adp}\uparrow$  & \textbf{$F_{\beta}^{adp} \uparrow$} &  $S\uparrow$ & $MAE\downarrow$ & $E_{\xi}^{adp}\uparrow$  & \textbf{$F_{\beta}^{adp} \uparrow$}  \\
			\midrule
			BBSNet \cite{fan2020bbs} & 90.67 & 0.46 & 86.07 & 67.12  & 90.27 & 0.41 & 86.87 & 68.43  & 92.74 & 0.46 & 91.91   & 77.23  & 85.48 & 0.28 & 65.83  & 41.83   \\
			DPANet \cite{chen2020dpanet} & 70.68 & 1.89 & 49.05  & 24.17   & 72.26 & 1.82 & 55.24   & 32.07  & 76.65 & 2.06 & 65.92  & 44.05  & 62.04 & 1.08 & 30.16   & 5.97 \\
			RD3D \cite{chen20223}   & 90.31 & 0.46 & 81.78 & 61.80 & 90.76 & 0.44 & 82.94  & 63.97  & 92.89 & 0.52 & 89.08  & 73.35  & 84.66 & 0.24 & 57.07  & 33.14  \\
			SwinNet(VD) \cite{liu2021swinnet} & {91.55} & 0.34  & 88.70  & 71.11 & 91.27 & 0.44 & 88.77 & 71.56 & 93.30 & 0.47 & 93.21  & 79.82  & 86.13 & 0.24 & 66.73 & 42.93   \\
			HRTrans(VD)  \cite{tang2022hrtransnet} & 90.92 & 0.29 & 95.74  & 80.87  & 90.81 & 0.32 & 95.84  & 80.91  & 93.03 & 0.36 & 97.37  & 86.49  & 84.63 & \textcolor{blue}{0.12} & 85.45  & 61.91 \\
			HWSI \cite{song2022novel} & \textcolor{blue}{92.86} & \textcolor{green}{0.23} & \textcolor{blue}{98.05}   & \textcolor{blue}{86.55} & \textcolor{green}{92.85} & \textcolor{blue}{0.29} & \textcolor{blue}{97.73} &  \textcolor{blue}{85.86} & \textcolor{blue}{94.17} & \textcolor{blue}{0.35} & \textcolor{blue}{98.43}  & \textcolor{blue}{89.20}  & \textcolor{green}{89.45} & \textcolor{green}{0.08} & \textcolor{green}{93.62} & \textcolor{green}{75.10}  \\
			MFFNet \cite{wan2023mffnet} & \textcolor{red}{93.64} & \textcolor{green}{0.23} & \textcolor{green}{98.14}  & \textcolor{green}{86.76} & \textcolor{red}{93.27} & \textcolor{green}{0.26} & \textcolor{green}{98.23}  & \textcolor{green}{86.63} & \textcolor{red}{94.61} & \textcolor{green}{0.32} & \textcolor{green}{98.73} & \textcolor{green}{89.94}  & \textcolor{red}{90.47} & \textcolor{red}{0.07} & \textcolor{blue}{93.51}   & \textcolor{blue}{74.25} \\
			\midrule
			\midrule
			Ours & \textcolor{green}{92.98} & \textcolor{red}{0.22} & \textcolor{red}{98.83}  & \textcolor{red}{89.59} & \textcolor{blue}{92.66} & \textcolor{red}{0.25} & \textcolor{red}{98.36}  & \textcolor{red}{89.11} & \textcolor{green}{94.18} & \textcolor{red}{0.29} & \textcolor{red}{98.86} & \textcolor{red}{91.93}  & \textcolor{blue}{88.97} & \textcolor{red}{0.07} & \textcolor{red}{97.13}   & \textcolor{red}{81.34}  \\
			\bottomrule
		\end{tabular}
	}
\end{table*}
\begin{table*}[htbp]
	\centering
	\caption{Quantitative results (\%) in T challenges. Here, ``$\uparrow$'' (``$\downarrow$'') means that the larger (smaller) the better. The best three results in each column are marked in \textcolor{red}{red}, \textcolor{green}{green}, and \textcolor{blue}{blue}, respectively. {\textbf{Note}: \textcolor{red}{Red} indicates the best performance in each metric.}}
	\label{cmp-T}
	\resizebox{\textwidth}{!}{%
		\begin{tabular}{l|cccc|cccc|cccc}
			\toprule
			\multirow{3}{*}{Method} & \multicolumn{4}{c|}{\textbf{T-Cr}} & \multicolumn{4}{c|}{\textbf{T-HR}} & \multicolumn{4}{c}{\textbf{T-RD}} \\
			\cmidrule(lr){2-5} \cmidrule(lr){6-9} \cmidrule(lr){10-13} 
			&  $S\uparrow$ & $MAE\downarrow$ & $E_{\xi}^{adp}\uparrow$  & \textbf{$F_{\beta}^{adp} \uparrow$} &  $S\uparrow$ & $MAE\downarrow$ & $E_{\xi}^{adp}\uparrow$  & \textbf{$F_{\beta}^{adp} \uparrow$} &  $S\uparrow$ & $MAE\downarrow$ & $E_{\xi}^{adp}\uparrow$  & \textbf{$F_{\beta}^{adp} \uparrow$}  \\
			\midrule
			CGFNet \cite{wang2021cgfnet} & 90.10 & 0.34 & 91.03  & 74.04  & 95.23 & 0.29 & 96.72  & 83.39  & 93.02 & 0.46 & 97.61  & 85.50 \\
			CSRNet \cite{li2021salient} & 84.53 & 0.50 & 92.75  & 73.69  & 93.74 & 0.33 & 98.12  & 89.14  & 89.81 & 0.61 & 97.76  & 84.97  \\
			DCNet \cite{tu2022weakly}  & 85.82 & 0.39 & 95.10  & 81.97  & 93.05 & 0.31 & 98.94  & 89.95 & 90.38 & 0.51 & 98.33  & 88.76  \\
			LSNet \cite{zhou2023lsnet}  & 88.08 & 0.42 & 91.37  & 73.25 & 90.95 & 0.43 & 95.67 & 79.60 & 90.11 & 0.64 & 96.97  & 81.73 \\
			SwinNet(VT) \cite{liu2021swinnet} & \textcolor{green}{92.63} & 0.26 & 92.53  & 77.18 & \textcolor{red}{96.08} & \textcolor{green}{0.21} & 97.69  & {85.98}  & \textcolor{red}{94.34} & \textcolor{green}{0.37} & 98.01  & 87.43  \\
			HRTrans(VT)  \cite{tang2022hrtransnet} & 91.30 & 0.27 & 95.42  & 81.13  & {94.82} & 0.22 & 98.67 & 87.88 & \textcolor{green}{93.92} & 0.38 & {98.69} & 88.47  \\
			HWSI \cite{song2022novel} & \textcolor{blue}{92.53} & \textcolor{green}{0.25} & \textcolor{blue}{97.44}  & \textcolor{green}{84.92}  & 94.70 & 0.26 & \textcolor{blue}{99.02}  & \textcolor{blue}{89.04}  & 93.24 & 0.40 & \textcolor{green}{99.01}  &  \textcolor{blue}{89.10}  \\
			MFFNet \cite{wan2023mffnet} & \textcolor{red}{92.87} & \textcolor{red}{0.23} & \textcolor{green}{97.49}  & \textcolor{blue}{84.90}  & \textcolor{green}{96.02} & \textcolor{red}{0.19} & \textcolor{green}{99.03} & \textcolor{green}{90.26} & {93.36} & \textcolor{green}{0.37} & \textcolor{blue}{98.95} & \textcolor{green}{89.82}  \\
			\midrule
			\midrule
			Ours & {91.90} & \textcolor{red}{0.23} & \textcolor{red}{97.85} & \textcolor{red}{87.91}  & \textcolor{blue}{95.29} & \textcolor{red}{0.19} & \textcolor{red}{99.43} & \textcolor{red}{92.20}  & \textcolor{blue}{93.70} & \textcolor{red}{0.35} & \textcolor{red}{99.22}  & \textcolor{red}{91.29} \\
			\bottomrule
		\end{tabular}
	}
\end{table*}

For more concrete analysis, we conduct a comparison against established multi-modal fusion methodologies across varying conditions, including adverse weather and partial sensor failure scenarios. As shown in Table \ref{tab:adverse_conditions}, single-modal and cross-modal fusion methods exhibits limitations in adverse scenarios due to less auxiliary modalities. Although multi-modal fusion methods, \emph{e.g.}, HRFuser \cite{broedermann2023hrfuser} and CMNeXt \cite{zhang2023delivering}, obtain good performance, our model is superior on most adverse scenarios thanks to the primitive fusion for multiple modalities. 
\subsubsection{VDT salient object detection}
To evaluate the effectiveness of our PWRF for VDT salient object detection, we compare it with 13 state-of-the-art approaches, including two RGB salient object detection methods (CPD \cite{wu2019cascaded} and RAS \cite{chen2018reverse}), three RGB-D salient object detection methods (BBSNet \cite{fan2020bbs}, DPANet \cite{chen2020dpanet}, and RD3D \cite{chen20223}), six RGB-T salient object detection methods (CGFNet \cite{wang2021cgfnet}, CSRNet \cite{li2021salient}, DCNet \cite{tu2022weakly}, and LSNet \cite{zhou2023lsnet}, SwinNet \cite{liu2021swinnet}, and HRTransNet \cite{liu2021swinnet}), and two VDT salient object detection methods (HWSI \cite{song2022novel} and MFFNet \cite{wan2023mffnet}). To ensure fair comparisons, all model predictions are either provided by the authors or generated using their source codes with default settings.

As shown in Table \ref{cmp-vdt}, three findings can be easily concluded: i) Compared with the RGB methods, most cross-modal methods achieve better performance, which is owing to the complementaries of depth and thermal modalities; ii) Compared with the V-D and V-T approaches, VDT methods both get higher metric values, which thanks to the complementary of triple modalities; iii) Compared with the previous VDT methods, our method achieves consistently superior performance for five best evaluation metrics, which owes to the superior multi-modal fusion of PWRF over the previous simple fusion mechanisms. To be more concrete, compared to the second-best model, MFFNet \cite{wan2023mffnet}, our model achieves significant improvements. Besides, our framework on cross-modal settings, including VD and VT, consistently achieves superior performance, which can be found from the second and third brackets in Table. \ref{cmp-vdt}, where our model on VD and VT conditions outperforms better than the other methods.

In addition, to demonstrate the robustness of our method in handling challenging scenarios, we present the performance comparison for visible-challenge, depth-challenge, and thermal-challenge scenes in Table \ref{cmp-V}, Table \ref{cmp-D}, and Table \ref{cmp-T}, respectively. For the challenging scenarios, our method still achieves a superior performance.
\subsection{Visual comparison}
\subsubsection{SMM semantic segmentation}
To visually demonstrate the performance of different models for SMM semantic segmentation, we selected two representative scenarios for comparison: ``Cloud \& Underexposure" and ``Rainy \& LiDAR Jitter". As shown in Fig. \ref{fig: visaul}\footnote{{The values of semantic labels are scaled via $ \times 10$ for visualization in Fig. \ref{fig: visaul}.}}, 
compared with the single-modal method SegFormer \cite{xie2021segformer}, multi-modal fusion methods provide more accurate semantic analysis. Moreover, compared with the multi-modal fusion method CMNeXt \cite{zhang2023delivering}, our method benefits from the Part-Whole Relational Fusion (PWRF) framework, resulting in more complete object segmentation in complex scenarios.
\begin{figure}[htbp]
	\centering
	\subfigure
	{
		\begin{minipage}[t]{0.86\linewidth}
			\centerline{\includegraphics[width=1\linewidth]{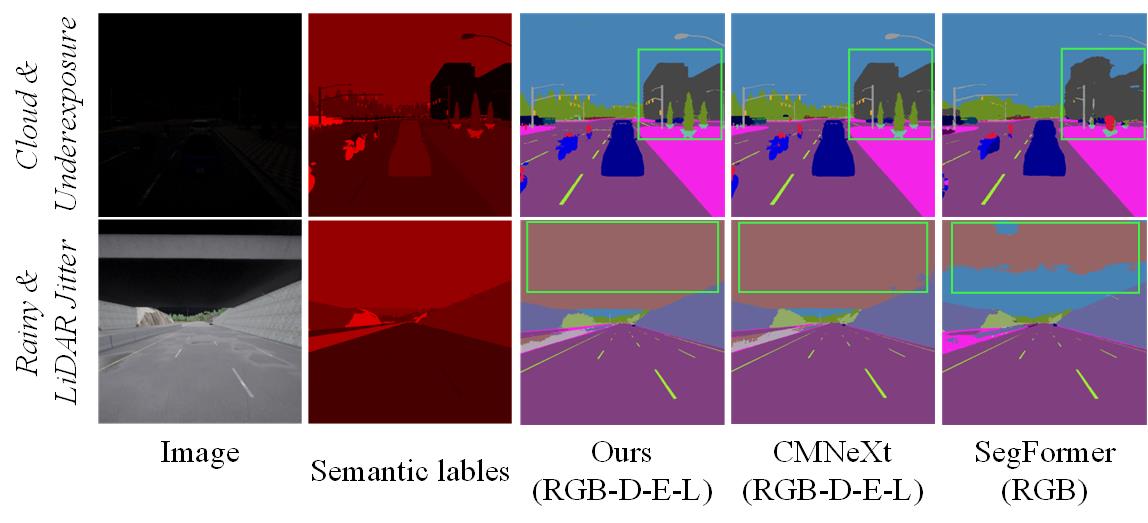}}
		\end{minipage}
		
	}
	\caption{{Visual comparison for SMM semantic segmentation.}}
	\label{fig: visaul}
\end{figure}
\subsubsection{VDT salient object detection}

To visually demonstrate the superiority of our model, we present several visualization results from challenging scenes across visible, depth, and thermal images. Specifically, V-challenges include big salient object (BSO), low illumination (LI), multiple salient objects (MSO), no illumination (NI), similar appearance (SA), side illumination (SI), and small salient object (SSO). D-challenges contain background interference (BI), background messy (BM), incomplete information (II), and small salient object (SSO). T-challenges cover crossover (Cr), heat reflection (HR), and radiation dispersion (RD). As shown in Fig. \ref{fig:visualVDT}(a), (b), and (c), the proposed model well tackles these challenging scenes for salient object detection, compared with the other methods.
\begin{figure}[htbp]
	\centering
	\subfigure
	{
		\begin{minipage}[t]{1\linewidth}
			\centerline{\includegraphics[width=1\linewidth]{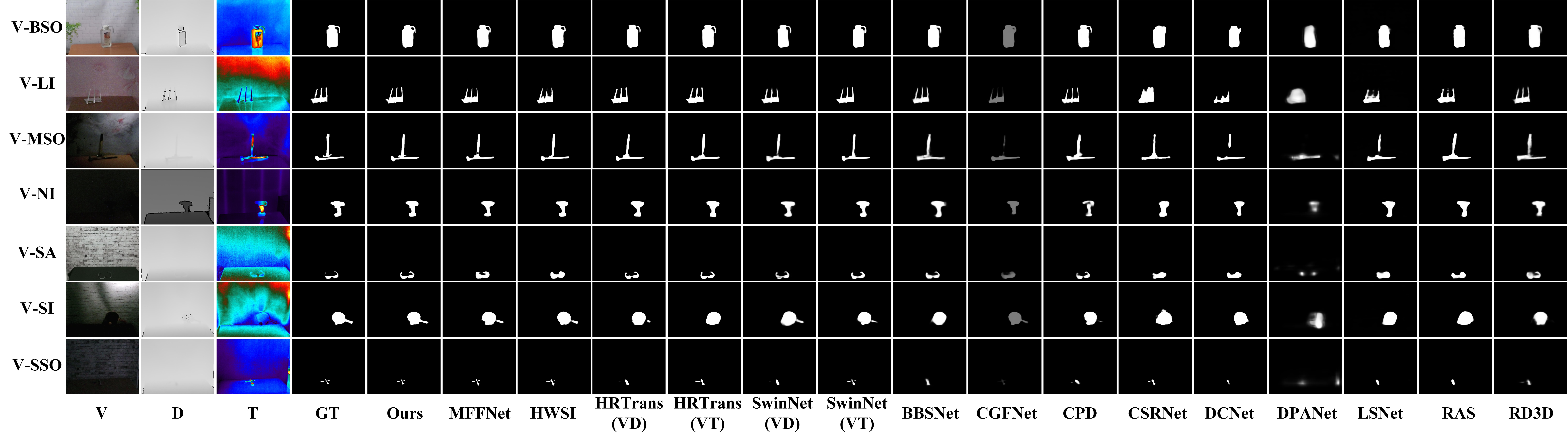}}
			\centerline{(a) Visual comparison of V-challenge.}
		\end{minipage}
		
	}
	\hfill
	\subfigure
	{
		\begin{minipage}[t]{1\linewidth}
			\centerline{\includegraphics[width=1\linewidth]{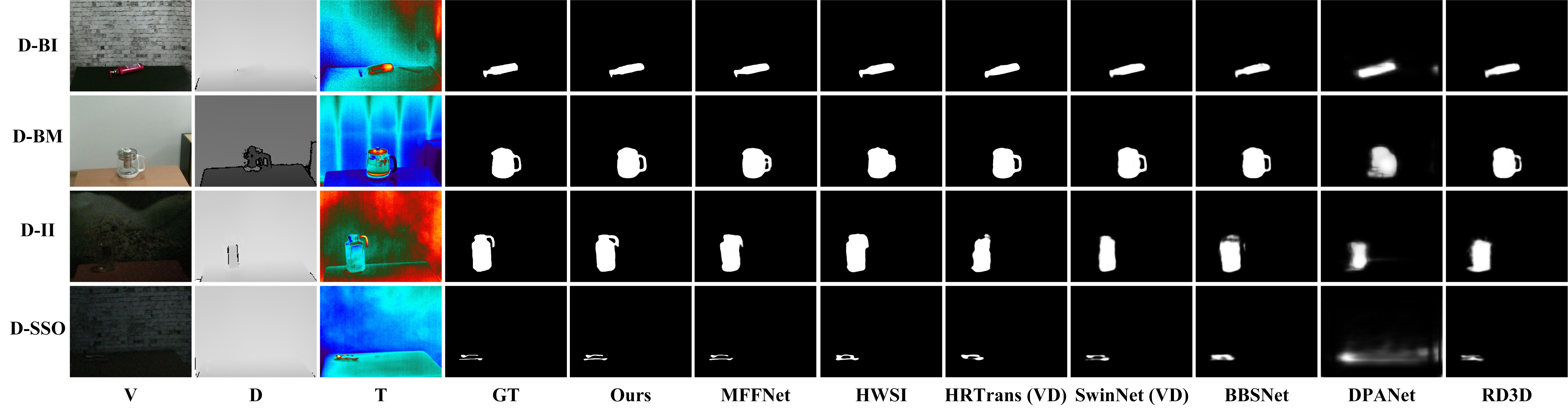}}
			\centerline{(b) Visual comparison of D-challenge.}
		\end{minipage}
	}
	\hfill
	\subfigure
	{
		\begin{minipage}[t]{1\linewidth}
			\centerline{\includegraphics[width=1\linewidth]{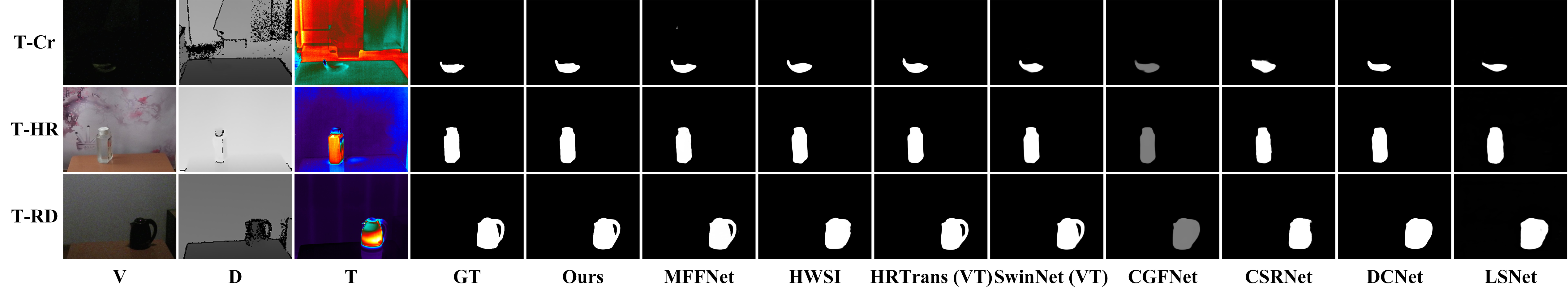}}
			\centerline{(c) Visual comparison of T-challenge.}
		\end{minipage}
	}
	\caption{Visual comparison of V-challenge, D-challenge, and T-challenge scenes for VDT salient object detection.}
	\label{fig:visualVDT}
\end{figure}
%
%
%
%
\subsection{Ablation study}
\subsubsection{Ablation analysis for SMM semantic segmentation}
In this subsection, we will conduct ablation experiments on DELIVER \cite{zhang2023delivering} to evaluate the contributions of key components of the SMM semantic segmentation model.

\textbf{Different number of capsule types for PWRF.}
Capsule type number takes a vital role in part-whole relations exploration, which will affect the performance of the whole model. To take a thorough study on different number of capsule types, we run several rounds with different part-level capsule types\footnote{The whole-level capsule types number is set to the category number for different datasets.}. As shown in Table \ref{n_of_cap}, it is seen that few or more capsules will lower the performance, because few capsules cannot find the accurate part-whole relations while more capsules will introduce noisy capsule assignments. By contrast, 8 types achieve the best IoU value, which is the setting for our model in this paper.

\textbf{Shared parameters.} Since there are multiple stages, our PWRF architecture should be repeated multiple times for different modality branch, which generates multi-branch structures. To discuss parameters sharing for these consistent structures, we carry out experiments using shared and unshared parameters for the model. As listed in the last two columns of Table \ref{n_of_cap}, shared parameters improve the semantic segmentation performance compared with the unshared setting. The reason behind comes from three folds: i) Shared parameters reduce some noise caused by the modality gap; ii) Shared parameters learn consistent fusion trend for different modalities; iii) By sharing structures and parameters, data from different modalities can assist each other to enhance the understanding of the same scenario.

\begin{table}[htbt]
	\centering
	\caption{Ablation study for different capsule types and parameters sharing on DeLIVER dataset \cite{zhang2023delivering}.}
	\label{n_of_cap}
	\begin{tabular}{c|c|c}
		\hline
		Primary Caps  & mIoU (Shared) & mIoU (Unsahred) \\
		\hline
		4 &   64.50   &  63.68\\
		8 &   \bf{66.47}  & \textbf{64.55} \\
		16 &  63.50   & 62.65 \\
		25 &  64.50  & 63.51 \\			
		\hline
	\end{tabular}
\end{table}

\subsubsection{Ablation analysis for VDT salient object detection}
In this subsection, we conduct ablation experiments on VDT-2048 dataset to evaluate the contributions of key modules in our proposed method.
\begin{table*}[htbp]
	\centering
	\caption{Ablation analysis (\%) on our baseline gradually including the newly proposed components on the VDT-2048 dataset \cite{song2022novel}.}
	\label{ablation_modules}
	\scalebox{0.82}{
		\begin{tabular}{c|c|c|c|c|c|c|c}
		\hline
		& Component & \textbf{$S \uparrow$} & \textbf{$MAE \downarrow$}  &\textbf{$E_{\xi}^{adp} \uparrow$} &\textbf{$E_{\xi}^{mean} \uparrow$}   & \textbf{$F_{\beta}^{adp} \uparrow$}  & \textbf{$F_{\beta}^{mean} \uparrow$}  \\
		\hline
		(a)     & Baseline  & 88.57 &    0.57    &   90.41  &  90.96  &   76.59  &   77.24    \\
		(b)     & + PWRF    & 92.24 &    0.28    &  98.37   &  98.31  &  87.35  &   88.37     \\	
		(c)     & + Stacking ASA deocder         & 90.14   &       0.41  &  94.91  &  95.48   &  84.03  &  84.36    \\
		(d)     & +PWRF + Stacking ASA deocder   & \bf{93.27} &  \bf{0.23}   &  \bf{98.84}  &  \bf{98.52}   & \bf{90.17} &   \bf{90.38}   \\
		
		\hline
	\end{tabular}
}
\end{table*}

\textbf{Different components.} To verify the effectiveness of different components in our proposed method, we perform various ablation experiments in Table \ref{ablation_modules}. First, comparing (a) \& (b) and (a) \& (c) in Table \ref{ablation_modules}, the proposed PWRF and stacking ASA decoder significantly boost the performance. The performance improvements come from two aspects: i) PWRF dynamically captures the informative semantics from different modalities for fusion; ii) Stacking ASA decoder helps to extract the primitive context of different modalities for prediction. Secondly, comparing (b) \& (d) and (c) \& (d) in Table \ref{ablation_modules}, the combination of PWRF and stacking ASA decoder achieves a higher performance, which proves the contributions of the proposed components efficiently. 

\textbf{Different fusion mechanisms.} In order to investigate the contribution of our PWRF for triple-modal fusion, we carry out several experiments by replacing our PWRF with different fusion mechanisms, including addition, concatenation, QKV attention mechanisms and EM routing \cite{hinton2018matrix} in our VDT salient object detection model. As shown in Table \ref{ablation_fusion_method}, there are two findings: i) Our PWRF surpasses the simple addition and concatenation mechanisms due to the primitive fusion for multiple modalities; ii) Compared with the attention mechanism, our model still achieves a significant superiority; iii) Our PWRF outperforms the previous EM routing \cite{hinton2018matrix} with a large margin, which demonstrates the superiority of DCR routing in our PWRF for VDT salient object detection.
\begin{table}[htbt]
	\centering
	\caption{Ablation study (\%) on different triple-modal fusion strategies on the VDT-2048 dataset \cite{song2022novel}.}
	\label{ablation_fusion_method}
	\begin{tabular}{c|c|c|c|c|c|c}
		\hline
		Settings & \textbf{$S \uparrow$} & \textbf{$MAE \downarrow$}  &\textbf{$E_{\xi}^{adp} \uparrow$} &\textbf{$E_{\xi}^{mean} \uparrow$}   & \textbf{$F_{\beta}^{adp} \uparrow$}  & \textbf{$F_{\beta}^{mean} \uparrow$}  \\
		\hline
		Addition                        &    91.72    &    0.32     &  95.96   &  96.82   &  85.98    &  87.14    \\
		Concatenation                   &    90.14    &    0.41     &  94.17   &  95.48   &  80.31    &  84.36    \\
		QKV Attention                       &    91.08    &    0.36     &  95.03   &  96.02   &  84.72    &  86.39    \\
		Concatenation + EM routing      &    71.59    &    1.04     &  82.57    &  87.13  &  33.88    &  52.67   \\
		\bf{Ours(PWRF)}                 & \bf{93.27}  &  \bf{0.23}  &  \bf{98.84}  &  \bf{98.52}   & \bf{90.17} &   \bf{90.38}    \\	
		
		\hline
	\end{tabular}
\end{table}

\textbf{Stacking ASA decoder.} Stacking ASA decoder contains two sub-decoders using a bridge connection. To deeply dig into its contribution, we conduct experiments including baseline by removing stacking ASA decoder from the entire model, one sub-decoder, and two sub-decoders. As shown in Table \ref{ablation_dad}, compared with baseline, one sub-decoder definitely improves the performance, which is because ASA emphasizes semantic feature channels while suppressing noisy ones. In addition, compared with one sub-decoder, stacking two sub-decoders performs better, which proves the bridge connection of two sub-decoders helps the model to get the superior performance.
\begin{table*}[htbt]
	\centering
	\caption{Ablation analysis (\%) for stacking ASA decoder on the VDT-2048 dataset \cite{song2022novel}.}
	\label{ablation_dad}
	\scalebox{0.78}{\begin{tabular}{c|c|c|c|c|c|c|c}
		\hline
		No & Settings & \textbf{$S \uparrow$} & \textbf{$MAE \downarrow$}  &\textbf{$E_{\xi}^{adp} \uparrow$} &\textbf{$E_{\xi}^{mean} \uparrow$}   & \textbf{$F_{\beta}^{adp} \uparrow$}  & \textbf{$F_{\beta}^{mean} \uparrow$}  \\
		\hline
		1     & Baseline + PWRF                      & 92.24        &       0.28    &  97.74  &  98.31    &   87.11   &   88.37     \\
		2     & Baseline + PWRF + one sub-decoder     & 92.47       &       0.26    & 97.92  &  98.33     &   87.95   &   88.73     \\	
		3     & Baseline + PWRF + two sub-decoders    & \bf{93.27}  &  \bf{0.23}  &  \bf{98.84}  &  \bf{98.52}   & \bf{90.17} &   \bf{90.38}     \\	
		\hline
	\end{tabular}}
\end{table*}

\textbf{Ablation analysis on different modalities.} To assess the impact of different modalities, we conduct four experiments detailed in Table \ref{ablation_modalities}, focusing on modalities combinations such as V+D, V+T, D+T, and V+D+T. Our PWRF method necessitates utilizing at least two distinct features from each modality, hence experiments on individual modalities were omitted. Moreover, for experiments involving combinations like V+D, V+T, and D+T, we simply removed one capsule feature branch while keeping other operations unchanged. From Table \ref{ablation_modalities}, it is evident that V+T obtains good performance compared with V+D and D+T. Leveraging three modalities prefer to improve the performance significantly, which demonstrates the superiority of more-modal fusion over cross-modal fusion.
\begin{table}[htbt]
	\centering
	\caption{Ablation study (\%) on different modalities on the VDT-2048 dataset \cite{song2022novel}.}
	\label{ablation_modalities}
	\begin{tabular}{c|c|c|c|c|c|c|c}
		\hline
		No & Settings & \textbf{$S \uparrow$} & \textbf{$MAE \downarrow$}  &\textbf{$E_{\xi}^{adp} \uparrow$} &\textbf{$E_{\xi}^{mean} \uparrow$}   & \textbf{$F_{\beta}^{adp} \uparrow$}  & \textbf{$F_{\beta}^{mean} \uparrow$}  \\
		\hline
		1     &V+D     & 90.84 &  0.33  &  97.57  &  97.61   &  84.77  &   85.51 \\
		2     &V+T     & 92.85 &  {0.26}  &  {98.65}  &  {98.43}   & {89.02}  &  89.45    \\
		3     &D+T     & 90.21  &  0.35   &   97.88  &  97.79      &   84.30   &   84.68    \\
		4     &V+D+T   & \bf{93.27}  &  \bf{0.23}  &  \bf{98.84}  &    \bf{98.52}   & \bf{90.17} &   \bf{90.38}   \\	
		
		\hline
	\end{tabular}
\end{table}
\subsubsection{Routing coefficients explanation}

Most previous methods cannot interpret the fusion of different modalities, which limits their reliability in real-world applications. In contrast, our model can provide an explanation for multi-modal fusion due to the part-whole relational routing from part-level modalities to whole-level modal. Specifically, we plot the horizontal and vertical routing coefficients for one pixel in Fig. \ref{fig:PF-POR}, in which higher routing values represent higher contribution of single modalities for fusion. {In Fig. \ref{fig:PF-POR}, we observe differences in the contributions from each modality under horizontal and vertical routing conditions. The x-axis, y-axis, and z-axis represent the part-level capsule types, whole-level capsule types, and routing coefficients, respectively.} For example, as shown in the red rectangle of Fig. \ref{fig:PF-POR}(a) in terms of the horizontal dimension, from the 8th part-level capsule to the first whole-level capsule, depth and event modalities contribute much for fusion, while LiDAR contributing less, which explains that depth and event modalities occupies much for semantic understanding while LiDAR has no roles at this pixel. {By contrast, in the vertical direction at the pixel, other modalities might dominate. Such differences could be due to the unique characteristics captured by each modality in different spatial dimensions. These directional differences in contribution help capture complementary information from different modalities, thus enhancing the feature fusion.}
\begin{figure}[htbp]
	\centering
	\subfigure
	{
		\begin{minipage}[t]{0.46\linewidth}
			\centerline{\includegraphics[width=1\linewidth]{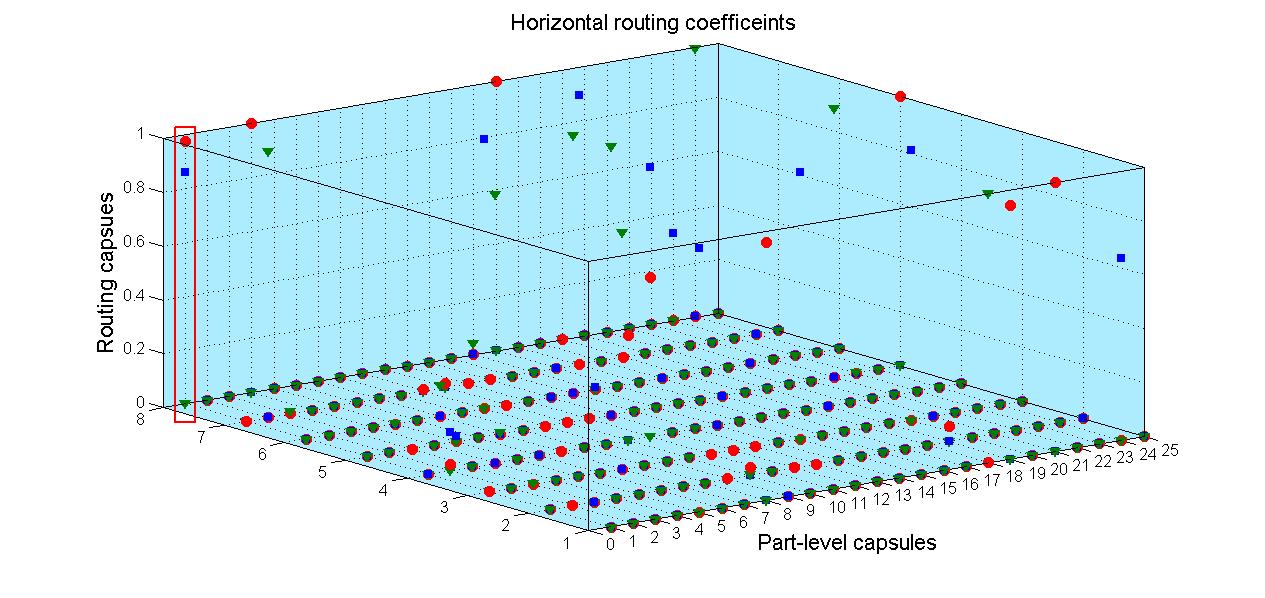}}
			\centerline{(a) Horizontal routing coefficients.}
		\end{minipage}
		
	}
	\hfill
	\subfigure
	{
		\begin{minipage}[t]{0.46\linewidth}
			\centerline{\includegraphics[width=1\linewidth]{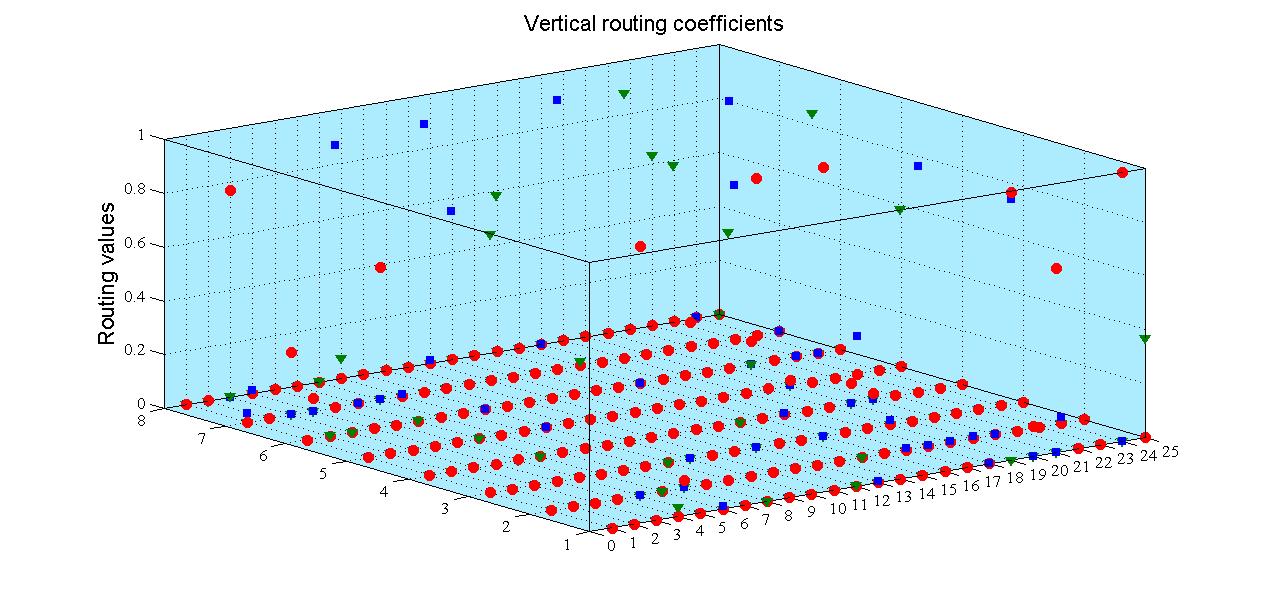}}
			\centerline{(b) Vertical routing coefficients.}
		\end{minipage}
	}
	\caption{Explanation using the routing coefficients. Red, blue, and green markers represent the routing coefficients of depth, event, and LiDAR modalities. x-axis, y-axis, and z-axis denote whole-level capsules types, part-level capsule types, and routing values, respectively.}
	\label{fig:PF-POR}
\end{figure}


\section{Limitations and Future Works}
\subsection{Limitations}
{\textbf{Complexity and Resource Requirements.} The proposed PWRF framework involves routing operations in CapsNets as well as multi-modal data fusion. Despite utilizing the lightweight DCR \cite{liu2022disentangled} mechanism, the framework still has high computational complexity and resource demands, especially when processing high-resolution and large-scale multi-modal data. This could limit its application in resource-constrained environments.}

{\textbf{Alignment and Noise in Multi-Modal Data.} Multi-modal sensors often suffer from spatial and temporal misalignment, and certain modalities (e.g., depth and event data) tend to contain significant noise. Although PWRF employs CapsNets to extract both modal-shared and modal-specific information, these issues have not been entirely addressed in the current implementation, which could negatively impact the overall quality of the results. Additionally, the current usage of attention mechanisms in the framework is relatively preliminary, and there is significant potential for further integration to improve feature selection and noise reduction, thereby enhancing robustness and the ability to capture critical information.}

{\textbf{Adaptability and Generalization.} While PWRF has demonstrated notable performance in applications such as autonomous driving and multi-modal object detection, its adaptability and generalization capabilities for other domains, such as multi-modal emotion recognition \cite{yang2023context} and medical imaging analysis \cite{rana2023machine}, have not been to be thoroughly evaluated. The effectiveness of the framework in these new domains remains an open problem for future research.}
\subsection{Future Work}
{\textbf{Lightweight for Real-Time Optimization.} We intend to further optimize the computational complexity of PWRF by exploring new lightweight  architectures or combining them with attention mechanisms \cite{alman2024fast,agarwal2023attention} to achieve higher performance in resource-limited environments, thus making the framework more suitable for real-time applications.}

{\textbf{Expansion to More Application Scenarios.} We intend to extend the PWRF framework to other multi-modal tasks, such as emotion recognition \cite{yang2023context} and medical imaging analysis \cite{rana2023machine}. Through experimentation in these new tasks, we hope to validate the generalizability of the model and improve its performance across different industries and applications.}

\section{Conclusion}
\label{sec:Conclusion}
In this article, we have presented a novel multi-modal fusion model from the perspective of part-whole relational fusion, which treated multi-modal fusion as routing each individual part-level modality to the fused whole-level modality. Using disentangled capsule routing, we modeled the modal-shared and modal-specific details for primitive fusion. Experiments on SMM semantic segmentation and VDT salient object detection demonstrate the superiority of the proposed PWRF framework for multi-modal scene understanding. In the future, we will study more primitive capsule routing for part-whole relational fusion and fusion explainability for reliable applications.

\bmhead{Acknowledgements}
This work is supported in part by the National Natural Science Foundation of Jiangsu Province under Grant No. BK20221379, in part by the State Key Laboratory of Reliability and Intelligence of Electrical Equipment under Grant EERI KF2022005, Hebei University of Technology.

\bibliography{sn-bibliography}

\end{document}